\documentclass[lettersize,journal]{IEEEtran}
\usepackage{amsmath,amsfonts}
\usepackage{algorithmic}
\usepackage{algorithm}
\usepackage{array}
\usepackage{textcomp}
\usepackage{stfloats}
\usepackage{url}
\usepackage{verbatim}
\usepackage{graphicx}
\usepackage{subfigure}
\usepackage{cite}
\usepackage{xcolor}

\usepackage{booktabs}
\usepackage{multirow}
\usepackage{makecell}
\usepackage{tabularx}
\usepackage{color, colortbl}
\definecolor{green_bcom}{RGB}{0, 205, 120}
\definecolor{yellow_bcom}{RGB}{255, 180, 0}
\definecolor{red_bcom}{RGB}{255, 80, 80}
\usepackage{tablefootnote}
\usepackage{adjustbox}
\usepackage[flushleft]{threeparttable}
\usepackage{hyperref}

\hypersetup{hypertex=true,
            colorlinks=true,
            linkcolor=blue,
            anchorcolor=blue,
            citecolor=blue}
            
\AtEndPreamble{
    \usepackage[capitalize]{cleveref}
    \crefname{section}{Section}{Secs.}
    \Crefname{section}{Section}{Sections}
    \crefname{table}{TABLE}{Tabs.}
    \Crefname{table}{TABLE}{Tables}
}

\def\BibTeX{{\rm B\kern-.05em{\sc i\kern-.025em b}\kern-.08em
    T\kern-.1667em\lower.7ex\hbox{E}\kern-.125emX}}
\begin{document}

\title{MUVOD: A Novel Multi-view Video Object Segmentation Dataset and A Benchmark for 3D Segmentation}

\author{Bangning Wei, Joshua Maraval, Meriem Outtas, Kidiyo Kpalma, Nicolas Ramin, Lu Zhang

\thanks{Bangning Wei is with the Institute of Research and Technology b$<>$com, Cesson-Sévigné, France and the Univ Rennes, INSA Rennes, CNRS, IETR - UMR 6164, F-35000 Rennes, France. (e-mail: bangning.wei@b-com.com)}

\thanks{Joshua Maraval and Nicolas Ramin are with the Institute of Research and Technology b$<>$com, Cesson-Sévigné, France. (e-mail: joshua.maraval@b-com.com; nicolas.ramin@b-com.com)}

\thanks{ Meriem Outtas, Kidiyo Kpalma, and Lu Zhang are with the Univ Rennes, INSA Rennes, CNRS, IETR - UMR 6164, F-35000 Rennes, France. (e-mail: meriem.outtas@insa-rennes.fr; kidiyo.kpalma@insa-rennes.fr; lu.zhang@insa-rennes.fr)}

\thanks{Corresponding author: Bangning Wei}

\thanks{\scriptsize
© 2025 IEEE. Personal use of this material is permitted. Permission from IEEE must be obtained for all other uses, in any current or future media, including reprinting/republishing this material for advertising or promotional purposes, creating new collective works, for resale or redistribution to servers or lists, or reuse of any copyrighted component of this work in other works.}}

\maketitle

\begin{abstract}
The application of methods based on Neural Radiance Fields (NeRF) and 3D Gaussian Splatting (3D GS) have steadily gained popularity in the field of 3D object segmentation in static scenes. These approaches demonstrate efficacy in a range of 3D scene understanding and editing tasks. Nevertheless, the 4D object segmentation of dynamic scenes remains an underexplored field due to the absence of a sufficiently extensive and accurately labelled multi-view video dataset. In this paper, we present MUVOD, a new multi-view video dataset for training and evaluating object segmentation in reconstructed real-world scenarios. The 17 selected scenes, describing various indoor or outdoor activities, are collected from different sources of datasets originating from various types of camera rigs. Each scene contains a minimum of 9 views and a maximum of 46 views. We provide 7830 RGB images (30 frames per video) with their corresponding segmentation mask in 4D motion, meaning that any object of interest in the scene could be tracked across temporal frames of a given view or across different views belonging to the same camera rig. This dataset, which contains 459 instances of 73 categories, is intended as a basic benchmark for the evaluation of multi-view video segmentation methods. We also present an evaluation metric and a baseline segmentation approach to encourage and evaluate progress in this evolving field. Additionally, we propose a new benchmark for 3D object segmentation task with a subset of annotated multi-view images selected from our MUVOD dataset. This subset contains 50 objects of different conditions in different scenarios, providing a more comprehensive analysis of state-of-the-art 3D object segmentation methods.  Our proposed MUVOD dataset is available at \url{https://volumetric-repository.labs.b-com.com/#/muvod}.
\end{abstract}


\section{Introduction}
\label{sec:intro}

\IEEEPARstart{T}{he} object segmentation of reconstructed real-world scenarios represents a challenging yet attractive topic in the fields of computer vision and graphics. The objective of this task is to extract information such as semantic classes or instance identities of target objects from the 3D representation of a scene, thereby enabling a more comprehensive understanding of the scene—an essential requirement for downstream tasks such as scene editing or object interaction in VR/AR applications \cite{chen20223d, piao2019real}. Traditional methods typically represent reconstructed scenes using explicit forms, such as voxel grids \cite{kar2017learning, penner2017soft} or textured meshes \cite{collet2015high, dou2016fusion4d, lv2021voxel}. A major limitation of these methods is their substantial data storage requirements, which present challenges for data compression and transmission. Moreover, purely explicit representations struggle to capture complex lighting effects, especially for non-Lambertian surfaces.

Neural Radiance Fields (NeRF) \cite{mildenhall2021nerf} has marked a significant advancement in 3D scene reconstruction, modeling scenes using learnable implicit functions trained on multi-view input images. Due to its cost-effective storage and ability to render photorealistic novel views, many NeRF-based approaches \cite{fridovich2022plenoxels, muller2022instant, barron2021mip, xu2022point, shen2023sd} have been proposed. More recently, 3D Gaussian Splatting (3D GS) \cite{ye2023gaussian} has attracted increasing attention from the research community. This method represents a 3D scene with explicit, three-dimensional Gaussians optimized using 2D images, and exhibits superior training and rendering speeds compared to NeRF-based methods.

The capacity of NeRF and 3D GS to accurately represent the photorealistic appearance of objects has led to a notable increase in the popularity of object segmentation methods \cite{goel2023interactive, cen2024segment, cen2023segment, ye2023gaussian, zhou2024feature} for static scenes based on these techniques. In contrast to traditional explicit segmentation approaches \cite{qi2017pointnet++, qiu2021geometric, wang20233d, du2024pcl}, these methods are trained on 2D object masks, which are easier to annotate for ground truth generation than 3D explicit representations.

Not limited to static scenes, dynamic scene reconstruction methods inspired by NeRF or 3D GS have been proposed at an accelerating pace \cite{song2023nerfplayer, liu2023robust, luiten2023dynamic, yang2024deformable, li2024spacetime, wu20244d}. The reconstructed 4D (3D + t) content representations are modeled from multi-view videos of dynamic scenes, thereby enabling users to freely change their viewing angles and navigate with six degrees of freedom (6DoF) in video sequences. Based on the principles of dynamic scene reconstruction, 4D object segmentation is a compelling research topic. It enables users to interact with objects during navigation in a way that is synchronized across both temporal and spatial axes. This represents a significant step forward in the field of virtual reality, with the potential to enhance the immersive experience of users and enrich the functionality of applications such as immersive learning \cite{marougkas2024personalized}.

In contrast to the numerous approaches for object segmentation in static 3D scenes, 4D object segmentation in dynamic scenes remains largely unexplored due to the lack of a richly annotated multi-view video segmentation dataset. While several multi-view video datasets provide ground truth masks, the available annotations are insufficient to support a thorough  understanding of complex, real-world scenarios. For example, Semantic-KITTI \cite{behley2019iccv} and KITTI-360 \cite{liao2022kitti} offer panoptic segmentation, but their classes are restricted to street objects and are predominantly used in the context of urban environments. IKEA ASM \cite{ben2021ikea} contains semantic labels only for furniture. Hi4D \cite{yin2023hi4d} and Lightfield \cite{mustafa20174d} annotate instance IDs for people, but their scenes contain at most two individuals. Human3.6M \cite{ionescu2013human3} provides bounding box annotations for people, while other datasets \cite{kim2012outdoor, zitnick2004high, ballan2010unstructured} offer only foreground silhouettes.

To support the development and evaluation of new methods, a dataset should contain videos from diverse real-life scenarios, comprising various object types and complex situations such as occlusion, in order to ensure robust evaluation \cite{ding2023mose}. For the 4D object segmentation task, each annotated object should have spatio-temporally coherent masks, meaning that the object can be consistently tracked across different reference views and temporal frames within a given video.

To address the issues discussed above, we present the Multi-view Video Object Segmentation Dataset (MUVOD). The dataset comprises 17 scenes collected from multiple sources, presented as sets of multi-view videos captured by camera rigs, and depicting a diverse range of activities including sports, urban environments, cooking, \emph{etc}. In each scene, three distinct camera views are selected for annotation, with each view containing 21 frames that have undergone manual refinement of the ground truth masks. Specifically, 459 instances across 73 categories are annotated in the panoptic format \cite{kim2020video} in multi-view videos, with each instance assigned both a semantic label and a unique instance ID, supporting robust 4D object tracking. Additionally, each object is assigned a motion status, selected from ``static", ``dynamic" and ``environmental", to enhance scene-level interpretation. We also provide a comparative analysis between MUVOD and existing multi-view video datasets with object-level annotations. To streamline the annotation process, we develop a spatio-temporal, semi-automatic method that substantially reduces manual workload while maintaining high annotation accuracy. Furthermore, we introduce a novel evaluation metric and a baseline approach, establishing MUVOD as a comprehensive benchmark for advancing research in multi-view video object segmentation.

Moreover, we introduce a new benchmark for the 3D object segmentation task, constructed using a subset of the MUVOD dataset. A total of 50 annotated instances are selected from diverse scenarios and categorized into four types based on their visual characteristics and contextual complexity: simple dominant objects, occluded objects, small-scale objects, and complex-structure objects. We evaluate the performance, robustness, and adaptability of four state-of-the-art 3D object segmentation methods on this benchmark and provide an in-depth analysis of their respective strengths and limitations. 

Our contributions are summarized as follows:
\begin{itemize}
    \item Development of the MUVOD dataset, enhancing research in 4D object segmentation in reconstructed real-world scenarios.
    \item Proposal of a semi-automatic annotation method for efficient production of ground truth segmentation masks.
    \item Creation of a novel task with an associated evaluation metric, designed to assess multi-view video object segmentation (VOS) methods.
    \item Evaluation of four state-of-the-art methods on a proposed 3D object segmentation benchmark containing multiple scenarios and different types of objects.
\end{itemize}

The remainder of this paper is organized as follows. \cref{sec:related_work} describes the related work. \cref{sec:proposed_dataset} presents the proposed MUVOD dataset and the annotation process. \cref{sec:problem definition} introduces the novel multi-view video object segmentation task, the evaluation metric and a baseline method. \cref{sec:3d_obj_seg} presents the new 3D object segmentation benchmark and experimental results of four algorithms. In \cref{sec:conclusion}, we conclude our paper.

\section{Related Work}
\label{sec:related_work}

\subsection{Scene Reconstruction from Multi-View Imaging}
Since the seminal introduction of NeRF, a neural volumetric rendering approach trained on multi-view images, the field has seen rapid and transformative progress. NeRF employs multilayer perceptron (MLP) parameters to implicitly model scenes and synthesize photorealistic views from novel perspectives. Its success in view synthesis has catalyzed the development of optimized variants aimed at improving reconstruction quality \cite{barron2021mip, barron2022mip, xu2022point, chen2022tensorf}. Among these, Plenoxels \cite{fridovich2022plenoxels} simplifies the plenoptic function using spherical harmonics on a voxel grid, significantly accelerating both rendering and training. Instant-NGP \cite{muller2022instant} introduces a hashing grid structure for efficient parameter indexing. Gaussian Splatting \cite{kerbl20233d}, by contrast, adopts an explicit representation based on 3D Gaussians whose parameters are optimized via a fast, differentiable rendering pipeline, allowing for more detailed and efficient scene capture. Collectively, these innovations have markedly improved the fidelity and computational efficiency of volumetric scene representations. As the field evolves, spatio-temporal extensions \cite{luiten2023dynamic} are emerging, promising even richer reconstructions of dynamic environments. In parallel, there is growing momentum behind advanced editing tools that enable interactive manipulation of these complex models \cite{garfield2024, NeRFshop23}, thereby expanding the horizons of digital scene reconstruction and immersive interaction.

\subsection{Radiance-based 3D Segmentation Benchmarks}
Inspired by NeRF and 3D Gaussian Splatting, numerous works \cite{zhi2021place, siddiqui2023panoptic, kundu2022panoptic} have integrated features extracted from multi-view 2D images into reconstruction pipelines to enable radiance-based 3D object segmentation. These methods are typically evaluated using existing multi-view image datasets, but the current benchmarks for this task remain scarce and limited in diversity. For instance, Replica \cite{straub2019replica} and Hypersim \cite{roberts2021hypersim} are indoor datasets offering multi-view imagery of furnished rooms, annotated with either semantic or panoptic masks. KITTI \cite{geiger2012we} and KITTI-360 \cite{liao2022kitti} provide street-level views captured from vehicle-mounted cameras, also with panoptic labels. While commonly used to train and assess 3D segmentation methods, these datasets are domain-specific and do not generalize well to a broader range of scenes. Several benchmarks exist, but they fall short in capturing the complexity of unconstrained environments. For example, NVOS \cite{ren2022neural}, based on LLFF \cite{mildenhall2019local}, includes 7 scenes, each with a single annotated target object visible only from one view. SPIn-NeRF \cite{mirzaei2023spin} provides 10 scenes, yet only a single isolated object is labeled in each. LERF-Mask \cite{ye2023gaussian} includes multiple objects, but covers just 3 indoor scenes.

In contrast, real-world environments typically involve multiple objects of varying scale, depth, and interaction. To resolve the issues with existing resources and enable fair evaluation of segmentation performance under more realistic conditions, we introduce a new benchmark derived from a subset of our MUVOD dataset.

\subsection{Video Segmentation}

Video segmentation methods are essential for understanding and analyzing dynamic scenes and can be broadly categorized into two main paradigms: video object segmentation and semantic segmentation-based approaches. Video object segmentation (VOS) methods \cite{wang2019fast, chen2020state, cheng2022xmem} focus on tracking a reference mask of a target object, typically provided in the first frame, and propagating it across the sequence. Techniques in this category leverage temporal coherence and motion estimation to preserve the object’s spatial consistency and identity throughout the video. VOS methods are frequently used to assist the annotation process, effectively reducing the manual burden on human annotators.

Semantic segmentation-based methods aim to provide a richer semantic explanation of video content. These include video semantic segmentation, instance segmentation, and panoptic segmentation. Video semantic segmentation \cite{gadde2017semantic} assigns a class label to every pixel, producing detailed maps of scene components. Video instance segmentation \cite{yang2019video} extends this by differentiating between individual instances of the same class, which is an essential capability in domains such as autonomous driving and surveillance. Video panoptic segmentation \cite{kim2020video, qiao2021vip} unifies both paradigms by assigning unique instance IDs to foreground ``thing" and semantic labels to background ``stuff". Recent progress in fine-grained segmentation \cite{wang2021amanet, wang2024cpi} further enhances this framework by capturing subtle structural variations, aiding in both object delineation and scene interpretation. These segmentation strategies, when applied in a multi-view context, hold strong potential for enhancing the semantic richness and spatial coherence of reconstructed scenes.

\subsection{Multi-view Video Datasets}

Multi-view video datasets aggregate dynamic scene recordings from multiple perspectives and are typically used for stereo or volumetric reconstruction. Some datasets acquire this information via a single moving camera \cite{yoon2020novel, luo2020consistent} or vehicle-mounted setups \cite{geiger2012we, liao2022kitti}, but these often lack temporal synchrony across viewpoints—a critical requirement for high-fidelity 4D reconstruction. For improved reconstruction quality, most contemporary datasets rely on camera rigs equipped with multiple calibrated and synchronized cameras to record multi-view videos. Many of these rig-based datasets focus on capturing human actions \cite{peng2021neural, fang2021mirrored, yin2023hi4d, kim2012outdoor, zitnick2004high, ballan2010unstructured, sigal2010humaneva, ionescu2013human3, mustafa20174d}, serving applications in human motion analysis and 4D body reconstruction \cite{mustafa2020semantically}. In contrast, some collections target scenes with multiple objects and complex backgrounds \cite{doyen2017light, tapie2021barn, tapie2021breakfast, salahieh2018kermit, mieloch2020mpeg, domanski2016multiview, sheng2021new, mieloch2023new, li2022neural, broxton2020immersive, hobloss2021multi, ben2021ikea}, offering more diverse content. However, these datasets often lack object-level annotations or provide only limited ones, making them insufficient for evaluating 4D object segmentation methods.

In summary, the development of robust segmentation approaches based on NeRF or 3D Gaussian Splatting is constrained by the lack of richly annotated, multi-object, multi-view datasets suited for complex dynamic scenes. These advanced techniques require high-quality supervision for both training and evaluation. To address this gap, we introduce MUVOD, a comprehensive multi-view video dataset featuring extensive object masks tailored for 4D object segmentation. Furthermore, we identify key limitations in existing 3D segmentation benchmarks, particularly in their scalability and reliability. To overcome these challenges, we propose a novel benchmark derived from MUVOD, establishing a more rigorous and representative standard for evaluating 3D object segmentation methods.

\section{Proposed Dataset}
\label{sec:proposed_dataset}
\subsection{Data Collection}

The scope of our work centers on 4D reconstruction and semantic analysis in the context of static camera rigs capturing dynamic, real-world scenarios. Within this setting, synthetic content or footage acquired via handheld or vehicle-mounted cameras is considered out of scope, as such data lacks the temporal alignment and spatial consistency required for high-quality reconstruction. To ensure meaningful and comprehensive evaluation, we assess the data diversity of our dataset along three key dimensions. The first dimension is scenario theme, which influences the semantic distribution of object categories. A dataset encompassing a wide range of thematic contexts enables a more accurate assessment of a method’s generalization ability. Second, we consider scene complexity, characterized by object interactions such as occlusion, disappearance, positional exchange, and other dynamic behaviors. These factors increase the difficulty of tracking and segmentation, thereby supporting more rigorous evaluation. Lastly, we account for camera rig configuration, which impacts the available viewpoints and tests the model’s ability to handle varying levels of view sparsity in scene reconstruction.

Using these criteria, we select 17 scenes from four sources, originally captured for dynamic view synthesis and immersive reconstruction. The videos, stored in YUV or MP4 formats and ranging from 65 to 300 frames, cover diverse subjects and rig setups (details are presented in \Cref{tab:summary}). Most scenes feature human-object interactions and realistic backgrounds, resulting in occlusions and layered dynamics. This makes the dataset well-suited for evaluating 4D segmentation, interaction, and editing methods.

\begin{table*}[htbp]
 \caption{Summary of the proposed dataset. The videos are selected from 4 different sources, captured by varied camera rig configurations. The objects are annotated in 3 different views with spatio-temporal cosistence. }

  \centering
  \small
  \begin{adjustbox}{max width=0.95\textwidth}
  \begin{threeparttable}
  
  \begin{tabular}{c *{4}{|c} |c}
    \toprule
    Scene & Rig arrangement &Theme & Views & Frames per view   & Source \\
    \midrule
    Painter\tnote{1} \cite{doyen2017light} & 4x4 planar &Painting & 16 & 300  & \multirow{10}{*}{MPEG}\\
    Breakfast \cite{tapie2021breakfast} & 5x3 planar &Daily & 15  & 97   & ~ \\
    Barn\tnote{2}  \cite{tapie2021barn}  & 5x3 planar &Working & 15  & 97   & ~ \\
    Frog \cite{salahieh2018kermit} & 13x1 line &Daily & 13   &300  & ~ \\
    Carpark \cite{mieloch2020mpeg} & 9x1 line &Street view & 9 & 250  & ~ \\
    PoznanStreet \cite{mieloch2020mpeg} & 9x1 line &Street view & 9  & 250  & ~ \\
    Fencing \cite{domanski2016multiview} & 10x1 linear arc &Sport & 10  & 250   & ~ \\
    CBABasketball\tnote{3}  \cite{sheng2021new} & 30x1 linear arc &Sport & 30 & 300  & ~  \\
    MartialArts \cite{mieloch2023new} & 2 vertically stacked arcs &Sport & 15  & 65    & ~ \\
    Blocks \cite{domanski2014poznan} & 10x1 linear arc  &Daily & 10   & 300  & ~\\
    \midrule
    MATF\tnote{4} \cite{hobloss2021multi}  & 10x1 stereo line &Film & 10  & 300   & \multirow{1}{*}IRT b$<>$com\\
    \midrule
    FlameSteak \cite{li2022neural}  & 2 vertically stacked arcs &Kitchen & 21  & 300   & \multirow{2}*{Meta}\\
    CoffeeMartini \cite{li2022neural} & 2 vertically stacked arcs &Kitchen & 18  & 300  &\\
    \hline
    AlexaMeadeExhibit \cite{broxton2020immersive} & Semi-sphere &Painting & 46  & 300   & \multirow{4}*{Google}\\
    AlexaMeadeFacePaint \cite{broxton2020immersive} & Semi-sphere &Painting & 46  & 300   &\\
    Dog \cite{broxton2020immersive} & Semi-sphere &Animal & 46  & 150   &\\
    Welder \cite{broxton2020immersive} & Semi-sphere &Working & 46 & 300   &\\
    \bottomrule
  \end{tabular}
  \begin{tablenotes}
    \item[1] Copyright: Technicolor - Armand Langlois. All rights reserved
    \item[2] Copyright: InterDigital. All rights reserved
    \item[3] Copyright: Alibaba Group Holding Limited. All rights reserved
    \item[4] Copyright © IRT b$<>$com 2021 – All Rights Reserved
  \end{tablenotes}
 \end{threeparttable}
 \end{adjustbox}
    
  \label{tab:summary}
\end{table*}
\subsection{Object Categories}
\label{subsec:object_categories}
The object categories in our dataset are based on VIPSeg \cite{miao2022large}, a large-scale classification scheme chosen for its diversity: covering 124 categories, including 58 ``thing" classes and 66 ``stuff” classes. To better align the taxonomy with the content of our scenes, we extend the label set by adding several new classes and introduce a revised labeling strategy. This is necessary because the original grouping in VIPSeg does not always reflect our use case. For instance, items like toy and book in the ``CoffeeMartini" scene are considered ``thing" classes in our dataset, whereas VIPSeg classifies them as ``stuff".

We divide objects into three groups: dynamic objects ($O_d$), static objects ($O_s$), and environmental objects ($O_e$). Dynamic objects are typically the central focus of the scene and exhibit continuous motion. Their segmentation masks often require frame-by-frame adjustment. Static objects remain motionless but can play important roles in the scene, often interacting with dynamic objects through occlusion or proximity. Environmental objects describe background elements or distant objects that receive less attention and typically do not influence the main scene dynamics. Both dynamic and static objects are treated as ``thing" classes. Each is assigned a unique instance ID that can be tracked across frames and across views from different cameras in the same rig. In contrast, environmental objects are treated as ``stuff" classes and are assigned only a semantic label and objects of the same class share the same instance ID. \cref{fig:statistic} presents the distribution of annotated object types across scenes.

\subsection{Annotation Method}
\begin{figure}[tb]
  \centering
  \tiny
  \includegraphics[width=\linewidth]{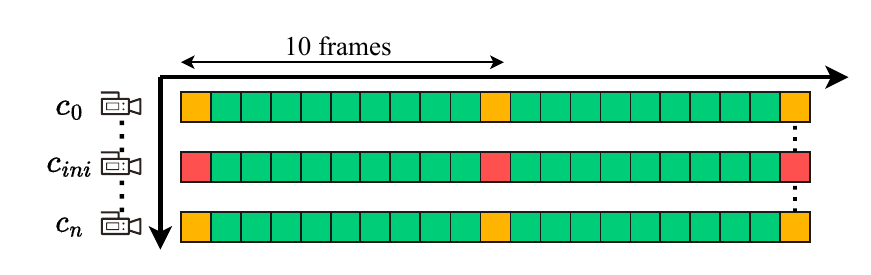}
  \caption{Annotation strategies. In our annotation work, all frames are classified into three types, representing different annotation strategies. The frames colored in red \fcolorbox{red_bcom}{red_bcom}{\rule{0pt}{2pt}\rule{2pt}{0pt}} are keyframes (chosen every 10 frames) of the video captured by the initial camera $c_{ini}$, which are manually annotated in the first stage. The frames in yellow \fcolorbox{yellow_bcom}{yellow_bcom}{\rule{0pt}{2pt}\rule{2pt}{0pt}} are keyframes of videos from other cameras, whose masks are propagated spatially from keyframes of $c_{ini}$ in the second stage. In green \fcolorbox{green_bcom}{green_bcom}{\rule{0pt}{2pt}\rule{2pt}{0pt}} are the other frames, the masks of which are propagated temporally from the corrected masks of keyframes in the corresponding video.
  }
  \label{fig:strategy}
\end{figure}

In an annotated multi-view video, object labels should remain consistent across frames and coherent across views. Manually annotating and associating the masks in such a video dataset is extremely time-consuming and expensive. Hence we introduce a semi-automatic annotation method to simplify and accelerate the annotation task. It consists of three steps: initial manual annotation, spatial propagation of labels from keyframes and bidirectional temporal mask propagation. The overall strategy is shown in \cref{fig:strategy}.

\begin{figure*}
  \centering
  \includegraphics[width=0.9\linewidth]{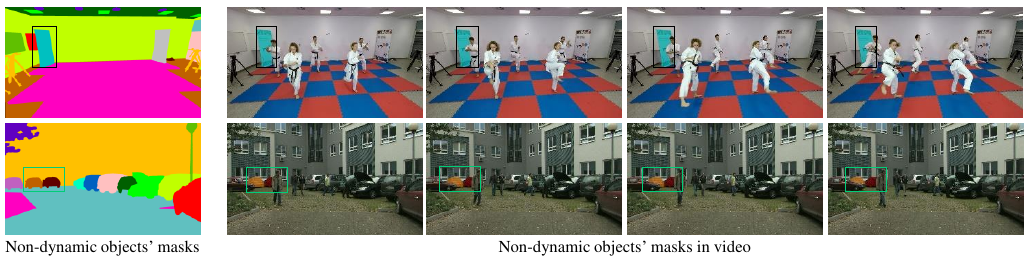}
  \caption{Examples of non-dynamic objects' masks. At initial keyframe annotation, non-dynamic objects are annotated as if there were no occlusion from dynamic objects. In parts of the video, these masks may be occluded by dynamic objects, such as the poster in the first line or the cars in the second line as depicted in the figure. The changes to these masks are automatically adjusted by superimposing masks of the nearer depth layer. }
  \label{fig:static_masks}
\end{figure*}

\textbf{Initial manual annotation:} For each scene, we select an initial camera $c_{ini}$ positioned near the center of the rig, providing comprehensive coverage of the dynamic objects and maximizing visibility of others. Keyframes are then sampled every 10 frames from the video captured by $c_{ini}$. In these keyframes, we manually annotate bounding boxes around the target objects. These boxes are then passed to the Segment Anything Model (SAM) \cite{kirillov2023segment}, a vision foundation model capable of generating high-quality object masks from bounding box prompts. The coarse masks generated by SAM are manually refined to ensure they are suitable for the subsequent spatial propagation step. Additionally, each object is assigned a depth layer to facilitate accurate placement of static and environmental elements. These objects are only annotated in the first keyframe and their masks are then propagated through the sequence (shown in \cref{fig:static_masks}). The depth layers allow us to treat occlusion deterministically: once the depth order is known, we can overlay masks from furthest to nearest, correcting only the dynamic object masks in later frames. This significantly reduces the need for frame-by-frame manual adjustments and efficiently resolves occlusion cases.

\textbf{Keyframes spatial propagation:} To maintain label coherence across different viewpoints, we treat the multi-view frames captured at a single time instant as a spatial video sequence. The annotated masks from the keyframes of the initial camera $c_{ini}$ are propagated to other camera views using the semi-supervised video object segmentation model XMem \cite{cheng2022xmem}. However, propagation quality may degrade depending on the camera rig structure, especially in configurations with wide spacing between views, such as double-arc or spherical rigs. To mitigate this, we incorporate 3D geometric cues to enhance object associations across views. We first identify the set of cameras most similar to $c_{ini}$ using a view similarity score, computed via view frustum intersection as proposed by Zamani \emph{et al}. \cite{zamani2017similarity}. Masks from $c_{ini}$ are propagated to these similar cameras. Then, we use LightGlue \cite{lindenberger2023lightglue} to extract keypoints from the high-confidence masks and triangulate them into 3D space. The resulting sparse 3D point clouds serve as pseudo ground truth, which we then project onto other camera views as spatial prompts for SAM. The masks generated by SAM from these projections help constrain the search space for XMem propagation, improving mask accuracy and reducing manual correction time. This integration of 3D guidance enhances cross-view consistency and annotation efficiency. 

\textbf{Bidirectional temporal propagation:} For each view, keyframe masks are manually corrected and then propagated in forward and backward frames. In contrast to conventional propagation in chronological order, this bidirectional operation is effective for obscured portions of objects not captured by the reference mask but revealed in subsequent frames.

Note that not all objects in the scene have been labelled, as some are too small or indistinguishable, and others are not relevant to the video's theme or do not interact with the main characters.

\subsection{Dataset Statistics}
\begin{figure*}
  \centering
  \includegraphics[width=0.85\linewidth]{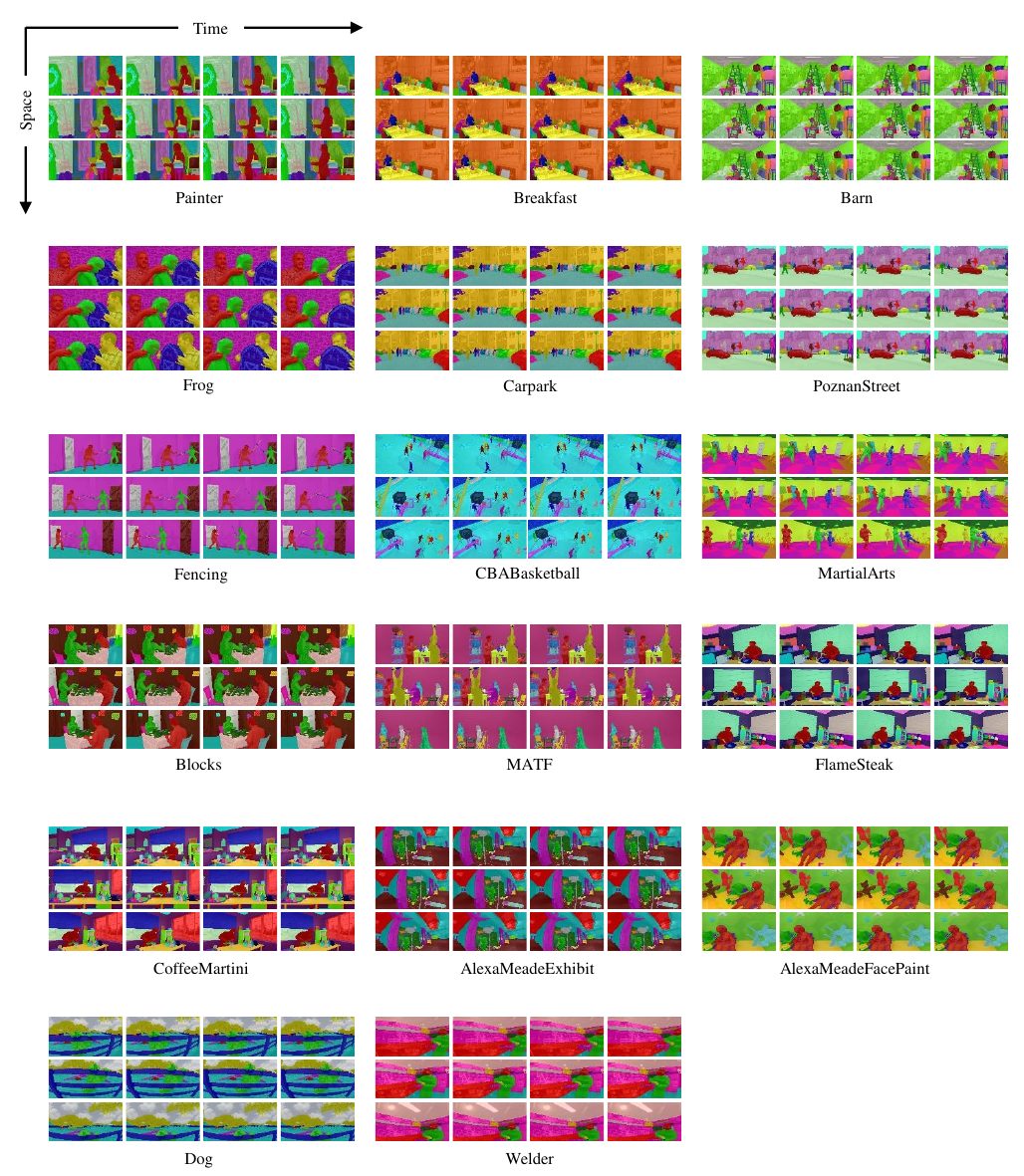}
  \caption{Visualization of the annotations of all the scenes.  For each scene, there are examples of videos taken by the initial camera (second row) and two other views. Each instance is distinguished by a distinct color within the scene. }
  \label{fig:example_anno}
\end{figure*}

\begin{figure*}[tb]
  \centering
  \includegraphics[width=0.8\linewidth]{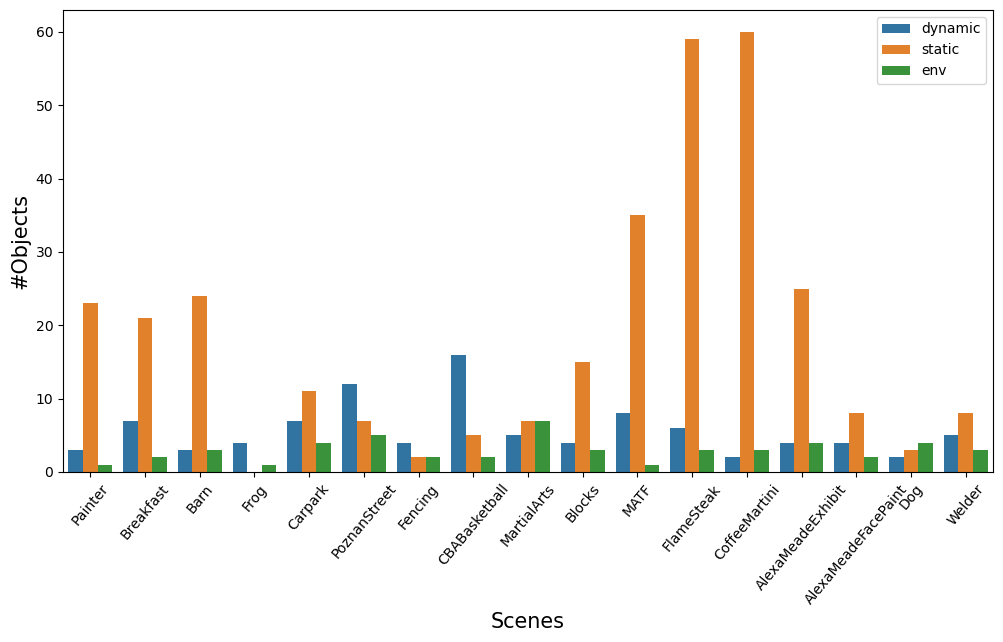}
  \caption{Number of different types of annotated objects in MUVOD scenes. The figure reflects the complexity of the scenes. ``CoffeeMartini" and ``FlameSteak" are two scenarios in a kitchen, containing 60 and 59 static objects, which are much more than other scenes. ``CBABasketball" has 16 dynamic objects, the most among the scenes. It captures a basketball game with many interactions between objects/person, which is challenging for a segmentation task.
  }
  \label{fig:statistic}
\end{figure*}

In total, we annotated 459 object instances of 73 categories within the 17 selected scenes. The number of different types of objects in each scene is shown in \cref{fig:statistic}. In most scenes, static objects outnumber both dynamic and environmental ones. Notably, the kitchen scenes ``CoffeeMartini" and ``FlameSteak" feature a dense collection of kitchenware, posing challenges in distinguishing and segmenting individual objects within visually similar categories. In contrast, scenes such as ``PoznanStreet", ``CBABasketball" and ``Frog" contain more dynamic objects than static ones, reflecting their emphasis on motion and activity. Annotation examples are visualized in \cref{fig:example_anno}.

\begin{figure}
  \centering
  \includegraphics[width=\linewidth]{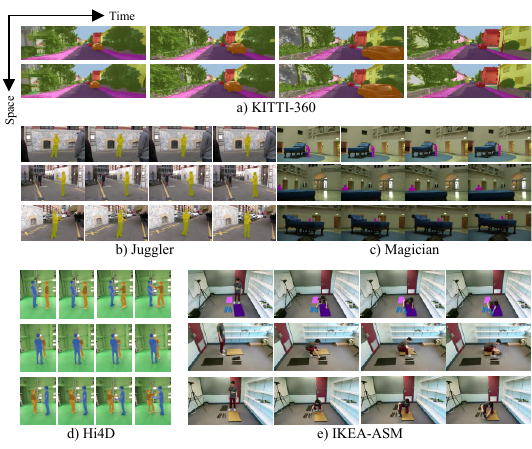}
  \caption{Visual examples of existing multi-view video datasets with annotations.
  }
  \label{fig:muvod_compare_4dseg}
\end{figure}

We compare our MUVOD dataset with other multi-view video datasets in \cref{tab:comparison}, based on five characteristics: number of viewpoints, camera type, scene theme, annotation type, and number of annotated classes, highlighting the advantages of our dataset. Semantic-KITTI \cite{behley2019iccv} and KITTI-360 \cite{liao2022kitti} offer panoptic segmentation, but their class coverage is limited to street-related objects. Both are captured using two moving cameras and are primarily designed for autonomous driving scenarios. IKEA ASM \cite{ben2021ikea} includes only three viewpoints and focuses solely on furniture, with annotations provided for a single view. Other datasets \cite{ionescu2013human3, mustafa20174d, yin2023hi4d, kim2012outdoor, zitnick2004high, ballan2010unstructured} may include more cameras, but their annotations are limited to foreground human subjects. These limitations make existing datasets unsuitable for evaluating 4D object segmentation, which requires a broader representation of object types and interactions in complex scenes. For visual comparison, annotation examples from several non-proprietary datasets are shown in \cref{fig:muvod_compare_4dseg}.

\begin{table*}

    \small
    \centering
    \caption{Comparison of existing multi-view video datasets with MUVOD }
    
    \begin{tabular}{c|ccccc}

    \hline
    Datasets & \#Views & Camera & Theme & Annotation & \#Classes  \\[1pt]
    \hline
    SemanticKITTI \cite{behley2019iccv} & 2 & dynamic & street view & panoptic & 28 \\[1pt]
    KITTI-360 \cite{liao2022kitti} & 2 & dynamic & street view & panoptic & 19  \\[1pt]
    IKEA ASM \cite{ben2021ikea} & 3 & static rig &  furniture assembly &  furniture part instance & 7 \\[1pt]
    Hi4d\cite{yin2023hi4d} & 8 & static rig &  human interaction & person instance & 2 \\[1pt]
    Handshake \cite{kim2012outdoor} & 8 & static rig &  human actions & silhouettes & 2 \\[1pt]
    Breakdance \cite{zitnick2004high} & 20 & static rig &  human actions & silhouettes & 2 \\[1pt]
    Magician \cite{ballan2010unstructured} & 5 & hand held &  human actions & silhouettes & 2 \\[1pt]
    Juggler \cite{ballan2010unstructured} & 6 & hand held &  human actions & silhouettes & 2 \\[1pt]
    HumanEva \cite{sigal2010humaneva} & 3 & static rig &  human actions & - & - \\[1pt]
    Human3.6M \cite{ionescu2013human3}  & 4 & static rig &  human actions & bounding boxes & 2 \\[1pt]
    Lightfield \cite{mustafa20174d}  & 20 & static rig &  human actions & person instance & 2 \\[1pt]
    \hline
    \textbf{Ours} & \textbf{9 to 46}  & \textbf{static rig} &  \textbf{diverse} & \textbf{panoptic} & \textbf{73} \\[1pt]
    \hline
    \end{tabular}
    
  \label{tab:comparison}
\end{table*}

\section{Multi-view Video Object Segmentation}
\label{sec:problem definition}

\subsection{Problem Definition}

 We introduce a new approach for video object segmentation titled ``Multi-view Video Object Segmentation". The purpose is to consistently segment a specific object or multiple objects from multi-view video sequences. This is a semi-supervised task, as the input is a ground truth mask from the original camera. Unlike conventional video object segmentation tasks, which focus mainly on challenges in the 2D realm, our proposed task introduces an extra layer of complexity. Operating within a multi-view context, our approach addresses issues related not only to temporal coherence but also spatial associations that arise from combining multiple viewpoints. This distinctive feature expands the scope of segmentation analysis, offering a more comprehensive evaluation framework for object segmentation methodologies.
 
\subsection{Evaluation Metric}
In the classical 2D semi-supervised video object segmentation task, the commonly used evaluation metric is the combination of the region similarity $\mathcal{J}$ and the contour accuracy $\mathcal{F}$. Given a predicted segmentation mask M and the corresponding ground truth mask G, the region similarity is defined as the Intersection over Union (IoU) between them:
\begin{equation}
    \mathcal{J} = \frac{\left | M \cap G  \right |}{\left | M \cup G  \right |}.
\end{equation}
The average of all objects' segmentation accuracy in the video is then calculated to measure the total $\mathcal{J}$. To measure the boundary quality of the predicted mask, we extract the closed contours $c(M)$ and $c(G)$ of the masks and compute the contour-based precision and recall $P_{c}$ and $R_{c}$ between the contour points via bipartite graph matching. The contour accuracy is defined as:
\begin{equation}
    \mathcal{F} = \frac{2P_{c}R_{c}}{P_{c} + R_{c}},
\end{equation}
Similarly to $\mathcal{J}$, we calculate the average of all objects' $\mathcal{F}$ as a total measure. Finally,
\begin{equation}
    \mathcal{J} \& \mathcal{F} = \frac{\mathcal{J} + \mathcal{F}}{2}
\end{equation}
is used to evaluate the overall segmentation accuracy of all objects through the sequence.

In our task, we extend the conventional 2D video evaluation metric to the multi-view scenario, aiming to assess the efficiency of a method in spatio-temporal mask propagation. Given a scene captured by a rig of $N$ cameras, an initial ground-truth mask for one keyframe captured by camera $c_{ini}$ is provided and masks for the rest of the frames should be propagated. The evaluation metric involves calculating the $\mathcal{J} \& \mathcal{F}$ between the predicted masks and the annotated ground-truth masks of videos captured by $N$ selected cameras. It is noted as:
\begin{equation}
    \mathcal{J} \& \mathcal{F}^{N} = \frac{\sum_{i=0}^{N-1}(\mathcal{J} \& \mathcal{F})_{c_{i}}}{N} .
\end{equation}

\subsection{Baseline Evaluation}
\label{subsec:baseline}

For future evaluation of methods on the multi-view video object segmentation task, a comparative baseline is necessary. Due to the absence of existing approaches specifically designed for this setting, we adopt a straightforward two-stage baseline by applying XMem \cite{cheng2022xmem} both spatially and temporally across multi-view video sequences. We regard it as the baseline method for the task and evaluate its performance across all scenes in our MUVOD dataset.

The evaluation is conducted on videos from three selected cameras: the initial camera $c_{ini}$ and two additional cameras that observe the primary content captured by $c_{ini}$ from distinct viewpoints. Since some annotated objects are visible only in non-initial views, we introduce two evaluation protocols to accommodate this variability: basic evaluation and complete evaluation.

\textbf{The basic evaluation} follows the notion of 2D video object segmentation. The scores are calculated only with the objects visible on the input reference frame, aiming to evaluate the model's ability to segment target objects in 4D space. The results of the evaluation are shown on \cref{tab:eval_basic}.  The baseline achieves an accuracy of 79.4\% $\mathcal{J} \& \mathcal{F}^{3}$ across all scenes, with the highest 92.3\% $\mathcal{J} \& \mathcal{F}^{3}$ on the scene ``Frog"  and the lowest 64.9\% $\mathcal{J} \& \mathcal{F}^{3}$ on ``AlexaMeadeFacePaint". Additionally, the method is evaluated separately for different object types, and it reaches 77.8\% $\mathcal{J} \& \mathcal{F}^{3}$ on dynamic objects, 79.3\% $\mathcal{J} \& \mathcal{F}^{3}$ on static objects and 83.1\% $\mathcal{J} \& \mathcal{F}^{3}$ on environmental objects.

\begin{table*}
    \centering
    \small
    \caption{Results of the basic evaluation of baseline method. $\mathcal{J}^{3}$ and $\mathcal{F}^{3}$ represent the mean region similarity and mean contour accuracy of the objects appearing in the scenes, captured by 3 cameras. $\mathcal{J} \& \mathcal{F}^{3}$ is the mean of $\mathcal{J}^{3}$ and $\mathcal{F}^{3}$. $\mathcal{J} \& \mathcal{F}^{3}_{dyn}$, $\mathcal{J} \& \mathcal{F}^{3}_{sta}$ and $\mathcal{J} \& \mathcal{F}^{3}_{env}$ denote $\mathcal{J} \& \mathcal{F}^{3}$ on dynamic objects, static objects and environmental objects, respectively.  }
    \setlength{\tabcolsep}{4pt}
    \begin{tabular}{c|ccc|ccc}

    \hline
     & $\mathcal{J} \& \mathcal{F}^{3}$(\%) & $\mathcal{J}^{3}$(\%) & $\mathcal{F}^{3}$(\%) & $\mathcal{J} \& \mathcal{F}^{3}_{dyn}$(\%) & $\mathcal{J} \& \mathcal{F}^{3}_{sta}$(\%) & $\mathcal{J} \& \mathcal{F}^{3}_{env}$(\%) \\
    
    \hline
    Painter & 82.2 & 77.8 & 86.7 & 73.7 & 83.7 & 78.4 \\
    Breakfast & 68.8 & 60.3 & 77.4 & 81.8 & 62.3 & 83.7 \\
    Barn & 77.3 & 69.0 & 85.7 & 65.6 & 78.3 & 80.9 \\
    Frog & 92.3 & 93.4 & 91.1 & 91.7 & - & 94.5 \\
    Carpark & 85.3 & 78.3 & 92.3 & 88.1 & 80.4 & 93.7 \\
    PoznanStreet & 80.2 & 73.2 & 87.2 & 79.8 & 83.1 & 77.5 \\
    Fencing & 85.7 & 80.6 & 90.8 & 75.2 & 95.2 & 97.1 \\
    CBABasketball & 66.4 & 59.7 & 73.1 & 66.8 & 71.9 & 67.5 \\
    MartialArts & 83.7 & 79.3 & 88.0 & 84.6 & 77.4 & 90.1 \\
    Blocks & 80.5 & 78.8 & 82.2 & 60.4 & 87.5 & 89.5 \\
    MATF & 69.1 & 61.2 & 76.9 & 73.0 & 66.5 & 90.1 \\
    FlameSteak & 84.8 & 78.3 & 91.4 & 88.0 & 85.4 & 80.3 \\
    CoffeeMartini & 84.3 & 78.1 & 90.5 & 92.2 & 85.1 & 70.3 \\
    AlexaMeadeExhibit & 82.8 & 71.9 & 93.7 & 78.7 & 82.1 & 92.4 \\
    AlexaMeadeFacePaint & 64.9 & 61.1 & 68.7 & 65.0 & 77.5 & 54.1 \\
    Dog & 75.5 &76.0 & 75.0 & 85.9 & 62.6 & 76.2 \\
    Welder & 85.2 & 81.7 & 88.7 & 71.5 & 90.3 & 95.8 \\
    \hline
    \textbf{Global} & 79.4 & 74.0 & 84.7 & 77.8 & 79.3 & 83.1\\
    \hline
    \end{tabular}
    
  \label{tab:eval_basic}
\end{table*}

\textbf{The complete evaluation} takes all the labelled objects into consideration when calculating the segmentation scores. Compared with the basic evaluation, this complete version introduces an additional challenge: determining whether the tested method is capable of providing a full understanding of a scene to detect and segment all the objects. The results of the evaluation are shown on \cref{tab:eval_complete} which indicates that the baseline achieves an overall accuracy of 75.6\% $\mathcal{J} \& \mathcal{F}^{3}$. The highest score is 92.3\% $\mathcal{J} \& \mathcal{F}^{3}$ on the scene ``Frog", which is the same as the basic evaluation. However, the lowest score is 59.3\% $\mathcal{J} \& \mathcal{F}^{3}$ on the scene of ``MATF". For the evaluation of different types of objects, the score for dynamic objects does not show a significant difference, however, the score for static objects drops by 7.1 and the score for environmental objects drops by 4.7.

As the baseline method is natively mask-propagation-based, it lacks the capability to fully understand a 4D scene by leveraging the multi-view information. This leads to the suboptimal results in scenes such as ``MATF" and ``Dog", in which multiple objects are positioned at a distance from the primary actors, preventing them from being captured by the central cameras.

\begin{table*}
    \centering
    \small
    \caption{Results of the complete evaluation of baseline method. $\mathcal{J}^{3}$ and $\mathcal{F}^{3}$ represent the mean region similarity and mean contour accuracy of the objects appearing in the scenes, captured by 3 cameras. $\mathcal{J} \& \mathcal{F}^{3}$ is the mean of $\mathcal{J}^{3}$ and $\mathcal{F}^{3}$. $\mathcal{J} \& \mathcal{F}^{3}_{dyn}$, $\mathcal{J} \& \mathcal{F}^{3}_{sta}$ and $\mathcal{J} \& \mathcal{F}^{3}_{env}$ denote $\mathcal{J} \& \mathcal{F}^{3}$ on dynamic objects, static objects and environmental objects, respectively.  }
    \setlength{\tabcolsep}{4pt}
    \begin{tabular}{c|ccc|ccc}

    \hline
     & $\mathcal{J} \& \mathcal{F}^{3}$(\%) & $\mathcal{J}^{3}$(\%) & $\mathcal{F}^{3}$(\%) & $\mathcal{J} \& \mathcal{F}^{3}_{dyn}$(\%) & $\mathcal{J} \& \mathcal{F}^{3}_{sta}$(\%) & $\mathcal{J} \& \mathcal{F}^{3}_{env}$(\%) \\
    
    \hline
    Painter & 76.1 & 71.9 & 80.2 & 73.7 & 76.2 & 78.4 \\
    Breakfast & 64.1 & 56.2 & 71.9 & 81.8 & 56.2 & 83.7 \\
    Barn & 77.3 & 69.0 & 85.6 & 65.6 & 78.3 & 80.9 \\
    Frog & 92.3 & 93.4 & 91.1 & 91.7 & - & 94.5 \\
    Carpark & 85.3 & 78.3 & 92.3 & 88.1 & 80.4 & 93.7 \\
    PoznanStreet & 80.2 & 73.2 & 87.2 & 79.8 & 83.1 & 77.5 \\
    Fencing & 85.7 & 80.6 & 90.8 & 75.2 & 95.2 & 97.1 \\
    CBABasketball & 65.6 & 59.0 & 72.2 & 65.1 & 71.9 & 67.5 \\
    MartialArts & 83.7 & 79.3 & 88.0 & 84.6 & 77.4 & 90.1 \\
    Blocks & 73.2 & 71.5 & 74.8 & 60.4 & 74.3 & 76.7 \\
    MATF & 59.3 & 52.4 & 66.1 & 73.0 & 53.6 & 90.1 \\
    FlameSteak & 70.6 & 65.2 & 76.0 & 88.0 & 73.0 & 34.4 \\
    CoffeeMartini & 76.7 & 71.0 & 82.3 & 92.2 & 75.9 & 50.2 \\
    AlexaMeadeExhibit & 82.8 & 71.9 & 93.7 & 78.7 & 82.1 & 92.4 \\
    AlexaMeadeFacePaint & 60.4 & 56.9 & 63.8 & 65.0 & 62.0 & 54.1 \\
    Dog & 67.2 & 67.5 & 66.8 & 85.9 & 31.3 & 76.2 \\
    Welder & 85.2 & 81.7 & 88.7 & 71.5 & 90.3 & 95.8 \\
    \hline
    \textbf{Global} & 75.6 & 70.5 & 80.7 & 77.7 & 72.2 & 78.4\\
    \hline
    \end{tabular}

  \label{tab:eval_complete}
\end{table*}

\subsection{Performance Analysis}\label{subsec:st_eval}

To better understand the characteristics of the dataset and to identify factors that future methods should account for, we conduct an in-depth analysis of the baseline’s performance from two perspectives: the number of objects in a scene and the variation in camera viewpoints across scenes. In this section, we report results based solely on the basic evaluation, in order to avoid bias introduced by objects that are only partially captured in non-initial views.

\begin{figure}
    \centering
    \subfigure[]{\includegraphics[width=0.24\textwidth]{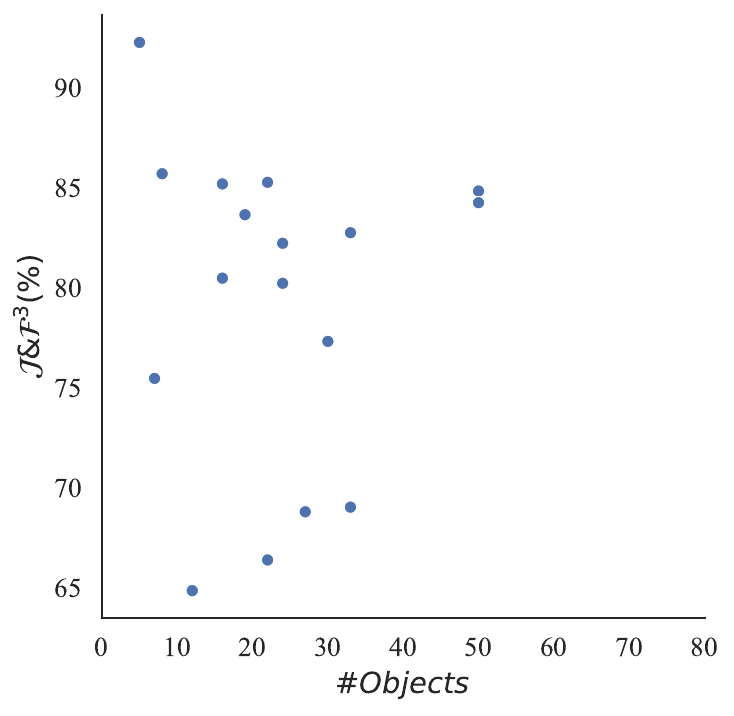}} 
    \subfigure[]{\includegraphics[width=0.24\textwidth]{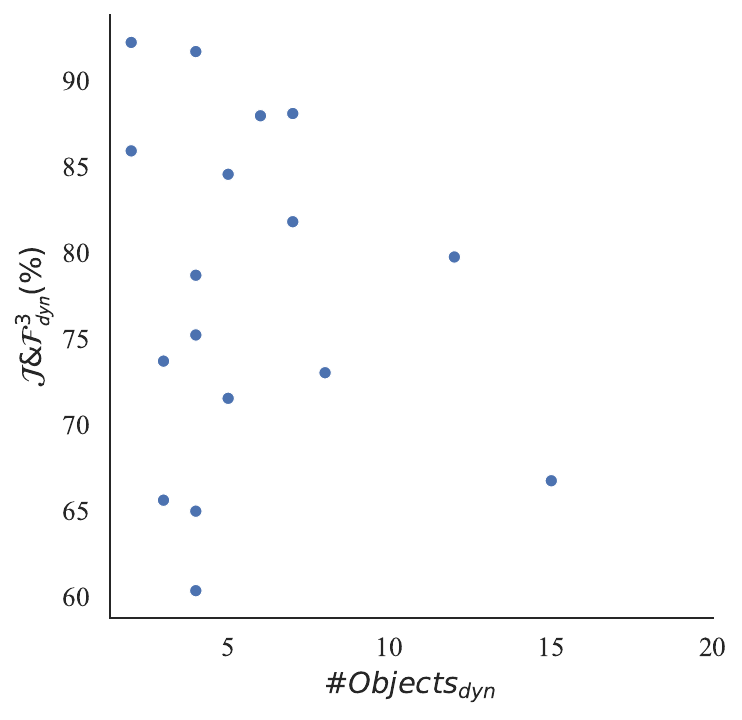}} 
    \subfigure[]{\includegraphics[width=0.24\textwidth]{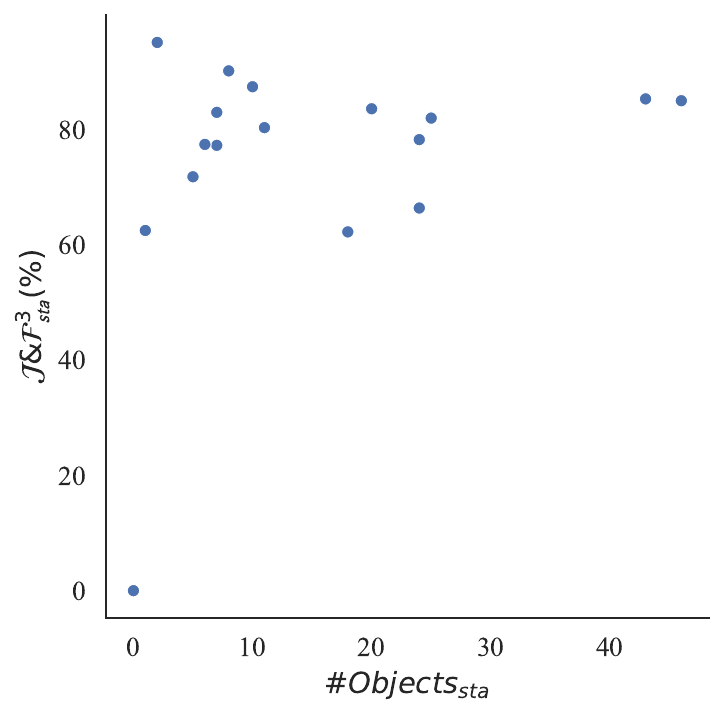}}
    \subfigure[]{\includegraphics[width=0.24\textwidth]{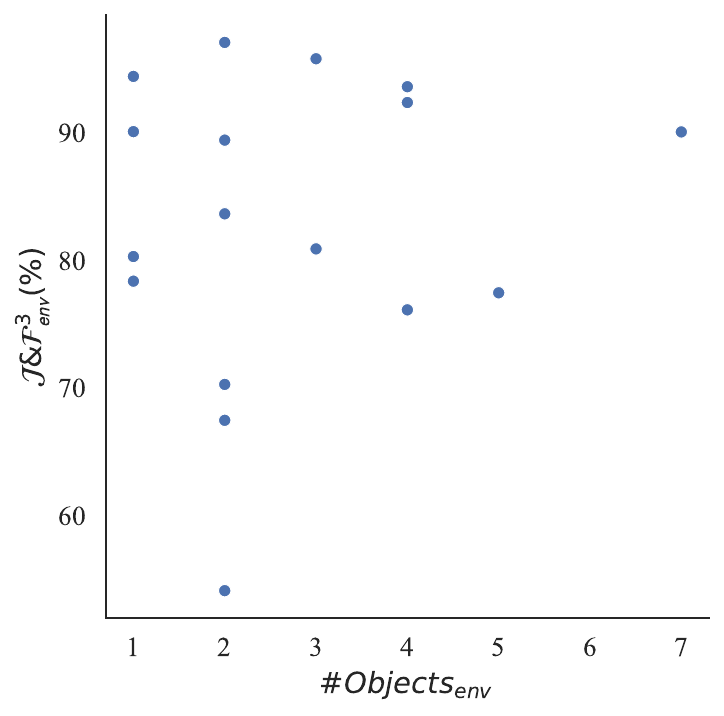}}
    \caption{The impact of object count in the scenes on the performance of the baseline model. The x-axis represents the number of objects in each scene, and the y-axis represents the $\mathcal{J} \& \mathcal{F}^{3}$ evaluated with the baseline model. Plot (a) considers all the annotated objects in each scene, plot (b) considers only the dynamic objects, plot (c) considers only the static objects and plot (d) considers only the environmental objects. }
    \label{fig:obj_num_baseline}
\end{figure}

\paragraph{Impact of object count}

\begin{figure}
  \centering
  \includegraphics[width=\linewidth]{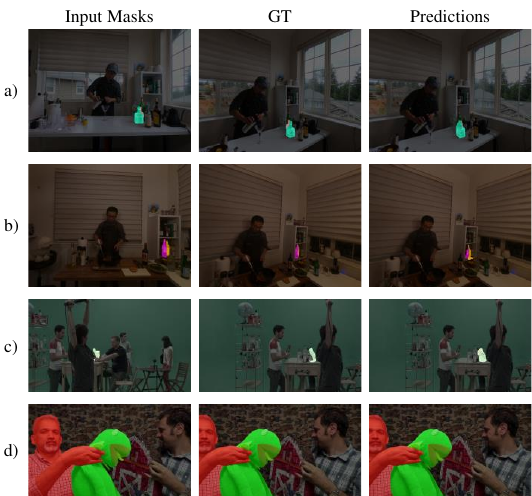}
  \caption{Examples of segmentation in scenes of different object counts. The sub-figures a), b) and c) show the results of scenes ``FlameSteak", ``CoffeeMartini" and ``MATF" which contain numerous objects in the scenes, increasing the possibility of erroneous segmentations on similar objects. The sub-figure d) shows the result of the scene ``Frog", which contains very few objects and the objects are large enough to be distinguished.
  }
  \label{fig:muvod_val_obj_count}
\end{figure}

Given the considerable variation in the number of annotated objects across scenes, we conducted an additional analysis to assess the impact of object count on the performance of the baseline model. The relationship between object count and evaluation score is visualized in \cref{fig:obj_num_baseline}, which breaks down the results by total objects, as well as dynamic, static, and environmental categories. Based on the point distribution, there is no strong correlation between the number of objects in a scene and the overall evaluation score. However, a larger number of objects increases the likelihood of segmentation errors in views that differ from the initial camera, particularly when objects exhibit similar textures or shapes. As shown in \cref{fig:muvod_val_obj_count}, objects in the ``bottle\_or\_cup" category are frequently missegmented, with masks incorrectly applied to nearby objects of the same class. In contrast, the ``Frog" scene, containing few, clearly distinguishable objects exhibits minimal segmentation errors. In real-world environments, increasing object count tends to raise scene complexity. Crowded scenes often lead to partial occlusions, reduced visibility, and visual overlap, which complicate object boundaries. Additionally, when objects vary in depth, smaller-scale instances become harder to distinguish, especially under low-contrast conditions. Although object count does not directly correlate with evaluation metrics, it implicitly affects segmentation difficulty by challenging the model’s ability to detect and delineate occluded or small-scale objects.

\paragraph{Impact of number of viewpoints} 

\begin{figure}
  \centering
  \includegraphics[width=\linewidth]{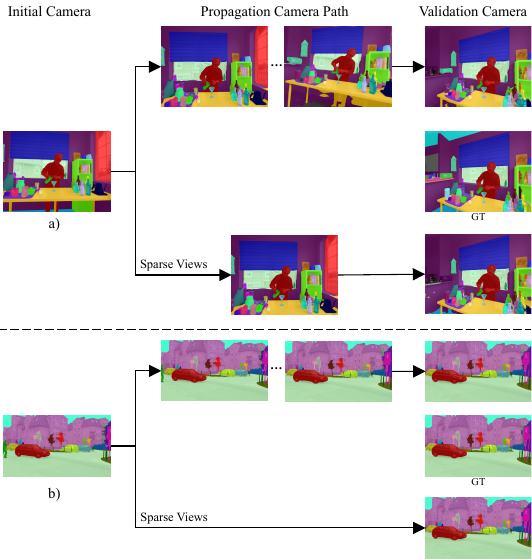}
  \caption{Visual examples of spatial propagation of full views and sparse views, where the scenes exhibit minimal discernible variation. The first line represents the propagation path of full views and the second line represents the propagation path of sparse views. The sub-figure a) shows the scene ``CoffeeMartini",  The sub-figure b) shows the scene ``PoznanStreet".
  }
  \label{fig:muvod_val_sparse_line}
\end{figure}

\begin{figure}
  \centering
  \includegraphics[width=\linewidth]{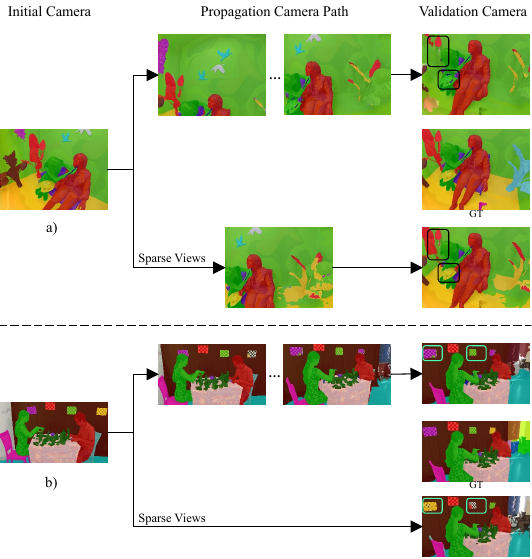}
  \caption{Visual examples of spatial propagation of full views and sparse views. The first line represents the propagation path of full views and the second line represents the propagation path of sparse views. The sub-figure a) shows the scene ``AlexaMeadeFacePaint", whose performance improves the most when fewer views are used. The sub-figure b) shows the scene ``Blocks", whose evaluation score decreases most. The objects have notable varying scorers are circled.
  }
  \label{fig:muvod_val_sparse}
\end{figure}

To examine the influence of viewpoint density on model robustness, we conduct an additional experiment by reserving only one-quarter of the available cameras and re-evaluating the baseline in this sparse-view setting. The results are presented in \Cref{tab:spa-tem-sparse}, where $\mathcal{J} \& \mathcal{F}^{3}$ denotes the original evaluation score, $\mathcal{J} \& \mathcal{F}^{3}_{sparse}$ reflects the performance under sparse conditions, and $Diff$ represents the score difference between the two settings. Some scenes show minimal performance change, primarily because they are captured using rig configurations with closely spaced cameras. In such cases, the reduction in available views has limited impact on the propagation process. Visual examples are shown in \cref{fig:muvod_val_sparse_line}. In contrast, other scenes exhibit notable score fluctuations, with some improving and others declining. To better understand these trends, we analyze the scenes with the most significant increases and decreases, as shown in \cref{fig:muvod_val_sparse}. For example, in ``AlexaMeadeFacePaint", performance improves under sparse-view evaluation. This is attributed to reduced view-to-view inconsistencies: when all views are used, certain objects intermittently disappear and reappear during mask propagation, leading to inaccuracies. Conversely, in ``Blocks", performance degrades due to insufficient camera coverage along the propagation path, resulting in inconsistent mask transfers.

These results suggest that both the number of annotated objects and the number of viewpoints significantly affect the performance of 4D object segmentation methods. The current baseline, which relies on 2D mask propagation, lacks integration of 3D spatial information and struggles with occluded or small-scale object segmentation. Furthermore, its dependency on consistent view-to-view transitions makes it sensitive to changes in camera configuration, leading to unstable propagation paths. As such, this baseline serves as a comparative starting point for future research. To advance this field, subsequent work should aim to develop more robust segmentation frameworks that incorporate 3D-aware representations and enable temporally consistent propagation of 3D masks, ultimately improving segmentation accuracy in 4D scenes.

\begin{table}
    \centering
    \footnotesize
    \caption{Results of the baseline method evaluated in sparse-view context. }

    \begin{tabular}{c|ccc}

    \hline
     & $\mathcal{J} \& \mathcal{F}^{3}$(\%) & $\mathcal{J} \& \mathcal{F}^{3}_{sparse}$(\%) & $ \left| Diff \right|$  \\
    
    \hline
    Painter             &  82.2 &   83.3 &      1.1 \\
    Breakfast           &  68.8 &   70.1 &      1.3 \\
    Barn                &  77.3 &   78.6 &      1.3 \\
    Frog                &  92.3 &   91.4 &      0.8 \\
    Carpark             &  85.3 &   84.4 &      0.8 \\
    PoznanStreet        &  80.2 &   80.0 &      0.2 \\
    Fencing             &  85.7 &   80.2 &      5.5 \\
    CBABasketball       &  66.4 &   61.3 &      5.1 \\
    MartialArts         &  83.7 &   81.5 &      2.1 \\
    Blocks              &  80.5 &   65.5 &     \textbf{15.0} \\
    MATF                &  69.1 &   64.7 &      4.3 \\
    FlameSteak          &  84.8 &   81.7 &      3.2 \\
    CoffeeMartini       &  84.3 &   84.8 &      0.5 \\
    AlexaMeadeExhibit   &  82.8 &   82.3 &      0.5 \\
    AlexaMeadeFacePaint &  64.9 &   78.1 &     \textbf{13.2} \\
    Dog                 &  75.5 &   79.9 &      4.4 \\
    Welder              &  85.2 &   83.8 &      1.4 \\               

    \hline
    \end{tabular}

  \label{tab:spa-tem-sparse}
\end{table}

\section{3D Object Segmentation}
\label{sec:3d_obj_seg}
3D object segmentation plays a fundamental role in understanding static scene representations and serves as a crucial first step for downstream tasks such as object editing and 3D scene composition. Moreover, it provides the basis for 4D object segmentation, which incorporates spatio-temporal dynamics in a “3D + t” framework. In this section, we introduce a 3D object segmentation benchmark built upon the diverse scenarios and rich instance annotations offered by our MUVOD dataset. We evaluate and compare several state-of-the-art methods quantitatively, aiming to provide a comprehensive analysis of their strengths and limitations.

\subsection{Benchmark}

Recent 3D segmentation methods lift 2D multi-view features \cite{goel2023interactive, tang2023scene} or masks \cite{ye2023gaussian, cen2024segment} into 3D space by learning radiance fields or reconstructing 3D Gaussian representations. These approaches enable interactive segmentation of target objects in 3D using minimal user input, such as points or strokes from a single or a few viewpoints, and have demonstrated impressive performance. However, the datasets used for their quantitative evaluation remain limited. NVOS \cite{ren2022neural}, built on LLFF \cite{mildenhall2019local}, includes only 7 scenes where each target object is centrally located and fully visible from all viewpoints at a close depth. SPIn-NeRF \cite{mirzaei2023spin} contains 10 scenes with isolated objects, but the camera viewpoints are closely spaced, resulting in minimal view variation. In contrast, real-world environments are significantly more complex, often involving multiple objects with varying scales, depths, and interactions. LeRF-Mask \cite{ye2023gaussian} addresses this to some extent by providing multi-object masks in realistic settings. However, it comprises only three indoor scenes, and the annotated objects are consistently placed on or around a table.

To evaluate the robustness and generalizability of existing methods under more diverse and challenging conditions, we construct a 3D object segmentation benchmark using a curated subset of our MUVOD dataset. Following the evaluation protocol used in NVOS and SPIn-NeRF, a user provides segmentation prompts (points or strokes) on a \textit{reference view} of a target object, and the model’s prediction on a \textit{target view} is evaluated using the Intersection over Union (IoU) with the corresponding ground truth mask. Unlike the isolated or single-scale objects in the previous dataset, our tested objects are in diverse conditions: 
\begin{itemize}
    \item[\textbullet] Simple dominant object is a dominant feature in the scene and is visible in all camera angles.
    \item[\textbullet] Occluded object is fully visible from some views but occluded from others. The captures of such an object from the reference camera and the target camera might differ in terms of visibility.
    \item[\textbullet] Small-scale object in the scene is much smaller than the main actors and may only be ignored by some of the cameras in the rig.
    \item[\textbullet] Complex-structure object often has a structure similar to skeleton or unevenly distributed parts, which can make segmentation difficult.
\end{itemize}

\begin{figure}[b]
  \centering
  \includegraphics[width=\linewidth]{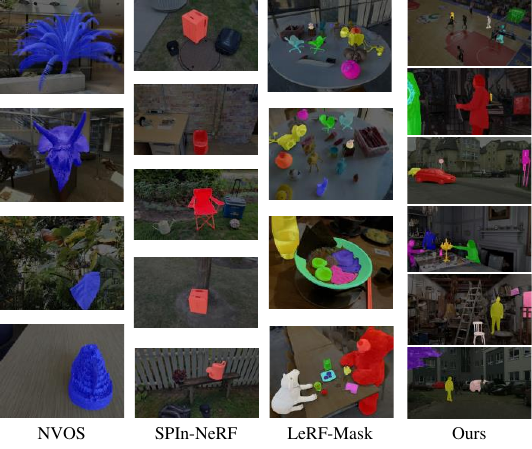}
  \caption{Visual examples of different 3D Object Segmentation datasets.
  }
  \label{fig:muvod_compare_3dseg}
\end{figure}

\subsection{Implementation Details and Results}

 To construct the experimental subset for 3D object segmentation, we select 12 scenes from the MUVOD dataset, each providing multi-view images of a single temporal frame. Four scenes captured using a semi-spherical rig are excluded due to significant image degradation caused by fisheye undistortion, which introduces artifacts and partial or fully cropped objects. These issues could lead to unfair comparisons during evaluation. Additionally, the ``MATF" scene is omitted because the reconstruction results using NeRF or 3D Gaussian Splatting are qualitatively suboptimal. In total, 50 target objects are chosen for segmentation, and their category distribution is presented in \Cref{tab:object_distribution}.

Compared with existing 3D object segmentation datasets (see \Cref{tab:compare_3dseg_dataset}), our benchmark includes a larger number of scenes, a broader range of thematic contexts, and more detailed annotations across a wider variety of object categories. These factors contribute to increased diversity and pose greater challenges to the robustness of evaluated methods. Visual examples of annotated instances are shown in \cref{fig:muvod_compare_3dseg}, where segmented objects are highlighted in distinct colors.

\begin{table*}
    \centering
    \small
    \caption{Distribution of tested target objects. }
    \begin{tabular}{c|c|c|c|c|c}

    \hline
    Datasets & Rig & Dominant & Occluded & Small-scale & Complex-structure  \\
    \hline
    Painter & 4x4 planar & 2 & 1 & 1 & 1 \\
    Breakfast & 5x3 planar &2 & 1 & 1 &1\\
    Barn & 5x3 planar &1 & 1 & 1 &1\\
    \hline
    Frog & 13x1 line &3 & 0 & 0 &1\\
    Carpark & 9x1 line &1 & 1 & 1 &1\\
    PoznanStreet & 9x1 line &1 & 1 & 1 &1\\
    \hline
    Fencing & 10x1 linear arc  &2 & 2 & 0 &0\\
    CBABasketball & 30x1 linear arc  &2 & 1 & 1 &0\\
    Blocks & 10x1 linear arc  &2 & 0 & 1 &1\\
    \hline
    MartialArts & 2 vertically stacked arcs  &2 & 2 & 0 &0\\
    FlameSteak & 2 vertically stacked arcs  &1 & 1 & 1 &1\\
    CoffeeMartini & 2 vertically stacked arcs  &1 & 1 & 1 &1\\
    \hline
    \textbf{Total} & - &\textbf{20}&\textbf{12}&\textbf{9}&\textbf{9} \\
    \hline
    \end{tabular}
    
  \label{tab:object_distribution}
\end{table*}

\begin{table}
    \setlength\tabcolsep{4pt}
    \renewcommand{\arraystretch}{1.1}
    \footnotesize
    \centering
    
    \caption{Comparison of 3D object segmentation benchmarks}

    \begin{tabular}{c|ccccc}

    \hline
    Dataset & Scenes & Labelled objects & Object type & Scene type  \\
    \hline
    NVOS \cite{ren2022neural} & 7 & 7 & Isolated & In/Outdoor \\
    SPIn-NeRF \cite{mirzaei2023spin} & 10 & 10 & Isolated & In/Outdoor  \\
    LERF-Mask \cite{ye2023gaussian} & 3 & 23 & Multiple & Indoor  \\
    \textbf{Ours} & \textbf{12} & \textbf{50} & \textbf{Multiple} & \textbf{Diverse}  \\
    \hline
    \end{tabular}
  \label{tab:compare_3dseg_dataset}
\end{table}

We evaluate four state-of-the-art 3D object segmentation methods: ISRF \cite{goel2023interactive}, SA3D \cite{cen2024segment}, SAGA \cite{cen2023segment}, and Gaussian Grouping \cite{ye2023gaussian} on our benchmark, using each method’s original prompting strategy. As shown in \cref{tab:quatiative_res}, Gaussian Grouping achieves the best performance with an average IoU of 78.8\%, showing strong adaptability to various object types. SA3D reaches 65.8\%, but struggles with occlusions and complex structures. SAGA and ISRF perform less effectively, scoring 47.2\% and 28.0\%, respectively.

\begin{table}[b]
    \setlength\tabcolsep{4.3pt}
    \renewcommand{\arraystretch}{1.3}
    \scriptsize
    \centering
    
    \caption{Quantitative results of 3D segmentation. Each column denotes the percent of mean Intersection over Union (mIoU\%) on different type of objects. One dominant object is selected from every scene, and each of the other three types contains 3 objects selected from different scenes.}

    \begin{tabular}{c|ccccc}

    \hline
    Methods & All & Dominant & Occluded & Small & Complex \\
    \hline
    ISRF\cite{goel2023interactive} & 28.0 & 38.8 & 24.0 & 15.2 & 22.1 \\
    SA3D\cite{cen2024segment} & 65.8 & 70.9 & 58.7 & 78.3 & 50.4 \\
    SAGA\cite{cen2023segment} & 47.2 & 63.7 & 44.3 & 23.8 & 37.0 \\
    Gaussian Grouping\cite{ye2023gaussian} & \textbf{78.8} & \textbf{84.5} & \textbf{69.4} & \textbf{86.4} & \textbf{71.1}\\
    \hline
    \end{tabular}
  \label{tab:quatiative_res}
\end{table}

\begin{figure}
  \centering
  \includegraphics[width=0.85\linewidth]{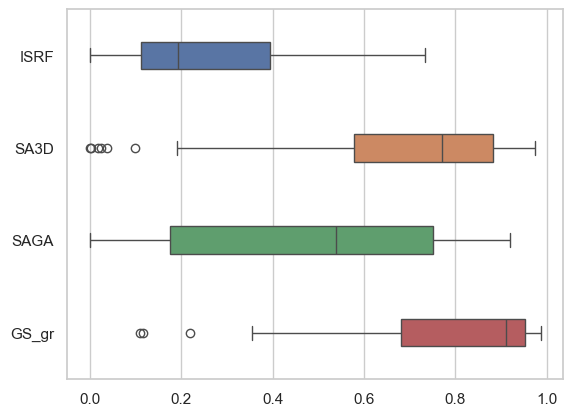}
  \caption{Statistical results of 3D object segmentation. The x-axis represents the IoU of different objects and the y-axis represents the evaluated methods (``GS\_gr" is the abbreviation for Gaussian Grouping). Statistical outliers are represented by circles. Although Gaussian Grouping and SA3D achieve acceptable average IoU, the performance varies greatly from object to object depending on the size of the object and its visibility in the scene.
  }
  \label{fig:3d_statistic}
\end{figure}

\subsection{Analysis and Discussion}
In this section, we propose an analysis of the results achieved by the methods and discuss the key to improving the performance of 3D object segmentation.

Gaussian Grouping is an effective 3D object segmentation approach. It treats multi-view images as video frames, segments objects using SAM, associates labels with a zero-shot video tracker DEVA \cite{cheng2023tracking}, and then jointly reconstructs an optimized 3D Gaussian Splatting (3DGS) model using these masks. The 3D Gaussians sharing the same label are grouped, naturally resulting in accurate projected masks. However, the method’s performance heavily depends on the quality and consistency of the initial masks across views. As illustrated in \cref{fig:3d_statistic}, Gaussian Grouping underperforms in several cases due to inaccurate initial segmentations or inconsistencies in object identity across viewpoints.

SA3D performs 3D segmentation via mask inverse rendering. It reprojects segmented pixels from a reference view into 3D space, renders them into novel views using a pretrained NeRF, and then uses these projections as prompts for SAM to generate confident masks. The final 3D mask is reconstructed by inverse rendering these masks back into 3D. This method is robust to object scale and scene diversity, and it rarely fails to localize the target. However, its accuracy is limited by the coarse quality of the rendered masks. Additionally, the time required to train NeRF and complete the reconstruction makes it unsuitable for real-time 4D applications.

SAGA also builds on SAM and 3DGS, but achieves only 47.2\% IoU, substantially lower than its reported results on simpler datasets like NVOS and SPIn-NeRF. It extracts object masks and features using SAM, and learns to associate objects across views by contrastively training the features of 3D Gaussians. While this is effective when the target object remains fully visible and relatively stable across views, it becomes unreliable in more complex scenes—especially for small-scale or distant objects, where appearance variation and occlusion hinder feature consistency.

\begin{figure*}[tb]
  \centering
  \includegraphics[width=\linewidth]{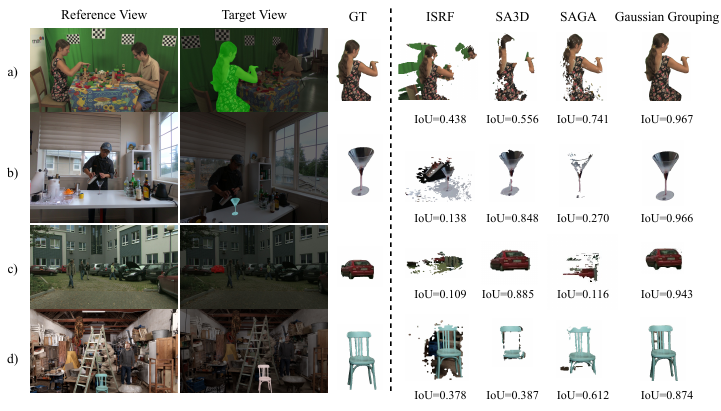}
  \caption{Comparison of the example ground truth (GT) objects with segmentation results of different methods: a) shows the dominant object b) shows the small-scale object. c) shows the occluded object. d) shows the complex-structure object.
  }
  \label{fig:subsets}
\end{figure*}

ISRF distills semantic features from the DINO ViT-b8 model \cite{caron2021emerging} into a voxelized radiance field. A seed is initialized by matching user-marked strokes to voxel features, and the 3D mask is expanded from this seed using bilateral search. The primary limitation of this method lies in its use of class-specific features, which makes it difficult to distinguish between different instances of the same semantic category. Moreover, the region-growing approach often fails to produce accurate masks in scenes with complex object interactions, such as the chair shown in ``example d)" of \cref{fig:subsets}.

Gaussian Grouping and SA3D achieve significantly better mean results, as presented in \Cref{tab:quatiative_res}, but the performance of all tested methods varies considerably depending on the category of the tested objects, sometimes failing entirely for outliers, as shown in \cref{fig:3d_statistic}.
While the benchmark subset of objects is small, it highlights the limitations of current methods in producing consistent 3D segmentation across diverse object classes.

\section{Conclusion}
\label{sec:conclusion}
This paper presents MUVOD, a new dataset for multi-view video object segmentation, comprising 17 real-world scenarios with rich annotations of 459 instances across 73 categories in a 3D + t context. We introduce a novel multi-view VOS task and an evaluation metric to facilitate future research on 4D object segmentation. Additionally, we conduct a 3D object segmentation benchmark on a representative subset of the dataset, evaluating four state-of-the-art methods and analyzing their strengths and limitations. Our findings underscore the importance of identity-consistent masks and optimized 3D representations. We hope MUVOD will serve as a valuable resource for advancing research in 4D scene segmentation, interaction, and editing.

\bibliography{references}

\begin{thebibliography}{10}
\providecommand{\url}[1]{#1}
\csname url@samestyle\endcsname
\providecommand{\newblock}{\relax}
\providecommand{\bibinfo}[2]{#2}
\providecommand{\BIBentrySTDinterwordspacing}{\spaceskip=0pt\relax}
\providecommand{\BIBentryALTinterwordstretchfactor}{4}
\providecommand{\BIBentryALTinterwordspacing}{\spaceskip=\fontdimen2\font plus
\BIBentryALTinterwordstretchfactor\fontdimen3\font minus \fontdimen4\font\relax}
\providecommand{\BIBforeignlanguage}[2]{{%
\expandafter\ifx\csname l@#1\endcsname\relax
\typeout{** WARNING: IEEEtran.bst: No hyphenation pattern has been}%
\typeout{** loaded for the language `#1'. Using the pattern for}%
\typeout{** the default language instead.}%
\else
\language=\csname l@#1\endcsname
\fi
#2}}
\providecommand{\BIBdecl}{\relax}
\BIBdecl

\bibitem{chen20223d}
S.-Y. Chen, Y.-K. Lai, S.~Xia, P.~L. Rosin, and L.~Gao, ``3d face reconstruction and gaze tracking in the hmd for virtual interaction,'' \emph{IEEE Transactions on Multimedia}, vol.~25, pp. 3166--3179, 2022.

\bibitem{piao2019real}
J.-C. Piao and S.-D. Kim, ``Real-time visual--inertial slam based on adaptive keyframe selection for mobile ar applications,'' \emph{IEEE Transactions on Multimedia}, vol.~21, no.~11, pp. 2827--2836, 2019.

\bibitem{kar2017learning}
A.~Kar, C.~H{\"a}ne, and J.~Malik, ``Learning a multi-view stereo machine,'' \emph{Advances in neural information processing systems}, vol.~30, 2017.

\bibitem{penner2017soft}
E.~Penner and L.~Zhang, ``Soft 3d reconstruction for view synthesis,'' \emph{ACM Transactions on Graphics (TOG)}, vol.~36, no.~6, pp. 1--11, 2017.

\bibitem{collet2015high}
A.~Collet, M.~Chuang, P.~Sweeney, D.~Gillett, D.~Evseev, D.~Calabrese, H.~Hoppe, A.~Kirk, and S.~Sullivan, ``High-quality streamable free-viewpoint video,'' \emph{ACM Transactions on Graphics (ToG)}, vol.~34, no.~4, pp. 1--13, 2015.

\bibitem{dou2016fusion4d}
M.~Dou, S.~Khamis, Y.~Degtyarev, P.~Davidson, S.~R. Fanello, A.~Kowdle, S.~O. Escolano, C.~Rhemann, D.~Kim, J.~Taylor \emph{et~al.}, ``Fusion4d: Real-time performance capture of challenging scenes,'' \emph{ACM Transactions on Graphics (ToG)}, vol.~35, no.~4, pp. 1--13, 2016.

\bibitem{lv2021voxel}
C.~Lv, W.~Lin, and B.~Zhao, ``Voxel structure-based mesh reconstruction from a 3d point cloud,'' \emph{IEEE Transactions on Multimedia}, vol.~24, pp. 1815--1829, 2021.

\bibitem{mildenhall2021nerf}
B.~Mildenhall, P.~P. Srinivasan, M.~Tancik, J.~T. Barron, R.~Ramamoorthi, and R.~Ng, ``Nerf: Representing scenes as neural radiance fields for view synthesis,'' \emph{Communications of the ACM}, vol.~65, no.~1, pp. 99--106, 2021.

\bibitem{fridovich2022plenoxels}
S.~Fridovich-Keil, A.~Yu, M.~Tancik, Q.~Chen, B.~Recht, and A.~Kanazawa, ``Plenoxels: Radiance fields without neural networks,'' in \emph{Proceedings of the IEEE/CVF Conference on Computer Vision and Pattern Recognition}, 2022, pp. 5501--5510.

\bibitem{muller2022instant}
T.~M{\"u}ller, A.~Evans, C.~Schied, and A.~Keller, ``Instant neural graphics primitives with a multiresolution hash encoding,'' \emph{ACM Transactions on Graphics (ToG)}, vol.~41, no.~4, pp. 1--15, 2022.

\bibitem{barron2021mip}
J.~T. Barron, B.~Mildenhall, M.~Tancik, P.~Hedman, R.~Martin-Brualla, and P.~P. Srinivasan, ``Mip-nerf: A multiscale representation for anti-aliasing neural radiance fields,'' in \emph{Proceedings of the IEEE/CVF International Conference on Computer Vision}, 2021, pp. 5855--5864.

\bibitem{xu2022point}
Q.~Xu, Z.~Xu, J.~Philip, S.~Bi, Z.~Shu, K.~Sunkavalli, and U.~Neumann, ``Point-nerf: Point-based neural radiance fields,'' in \emph{Proceedings of the IEEE/CVF Conference on Computer Vision and Pattern Recognition}, 2022, pp. 5438--5448.

\bibitem{shen2023sd}
S.~Shen, W.~Li, X.~Huang, Z.~Zhu, J.~Zhou, and J.~Lu, ``Sd-nerf: Towards lifelike talking head animation via spatially-adaptive dual-driven nerfs,'' \emph{IEEE Transactions on Multimedia}, 2023.

\bibitem{ye2023gaussian}
M.~Ye, M.~Danelljan, F.~Yu, and L.~Ke, ``Gaussian grouping: Segment and edit anything in 3d scenes,'' \emph{arXiv preprint arXiv:2312.00732}, 2023.

\bibitem{goel2023interactive}
R.~Goel, D.~Sirikonda, S.~Saini, and P.~Narayanan, ``Interactive segmentation of radiance fields,'' in \emph{Proceedings of the IEEE/CVF Conference on Computer Vision and Pattern Recognition}, 2023, pp. 4201--4211.

\bibitem{cen2024segment}
J.~Cen, Z.~Zhou, J.~Fang, W.~Shen, L.~Xie, D.~Jiang, X.~Zhang, Q.~Tian \emph{et~al.}, ``Segment anything in 3d with nerfs,'' \emph{Advances in Neural Information Processing Systems}, vol.~36, 2024.

\bibitem{cen2023segment}
J.~Cen, J.~Fang, C.~Yang, L.~Xie, X.~Zhang, W.~Shen, and Q.~Tian, ``Segment any 3d gaussians,'' \emph{arXiv preprint arXiv:2312.00860}, 2023.

\bibitem{zhou2024feature}
S.~Zhou, H.~Chang, S.~Jiang, Z.~Fan, Z.~Zhu, D.~Xu, P.~Chari, S.~You, Z.~Wang, and A.~Kadambi, ``Feature 3dgs: Supercharging 3d gaussian splatting to enable distilled feature fields,'' in \emph{Proceedings of the IEEE/CVF Conference on Computer Vision and Pattern Recognition}, 2024, pp. 21\,676--21\,685.

\bibitem{qi2017pointnet++}
C.~R. Qi, L.~Yi, H.~Su, and L.~J. Guibas, ``Pointnet++: Deep hierarchical feature learning on point sets in a metric space,'' \emph{Advances in neural information processing systems}, vol.~30, 2017.

\bibitem{qiu2021geometric}
S.~Qiu, S.~Anwar, and N.~Barnes, ``Geometric back-projection network for point cloud classification,'' \emph{IEEE Transactions on Multimedia}, vol.~24, pp. 1943--1955, 2021.

\bibitem{wang20233d}
Q.~Wang, D.~Liu, Z.~Liu, J.~Xu, and J.~Tan, ``3d object segmentation using cross-window point transformer with latent semantic boundary guidance,'' \emph{IEEE Transactions on Multimedia}, 2023.

\bibitem{du2024pcl}
A.~Du, T.~Zhou, S.~Pang, Q.~Wu, and J.~Zhang, ``Pcl: Point contrast and labeling for weakly supervised point cloud semantic segmentation,'' \emph{IEEE Transactions on Multimedia}, 2024.

\bibitem{song2023nerfplayer}
L.~Song, A.~Chen, Z.~Li, Z.~Chen, L.~Chen, J.~Yuan, Y.~Xu, and A.~Geiger, ``Nerfplayer: A streamable dynamic scene representation with decomposed neural radiance fields,'' \emph{IEEE Transactions on Visualization and Computer Graphics}, vol.~29, no.~5, pp. 2732--2742, 2023.

\bibitem{liu2023robust}
Y.-L. Liu, C.~Gao, A.~Meuleman, H.-Y. Tseng, A.~Saraf, C.~Kim, Y.-Y. Chuang, J.~Kopf, and J.-B. Huang, ``Robust dynamic radiance fields,'' in \emph{Proceedings of the IEEE/CVF Conference on Computer Vision and Pattern Recognition}, 2023, pp. 13--23.

\bibitem{luiten2023dynamic}
J.~Luiten, G.~Kopanas, B.~Leibe, and D.~Ramanan, ``Dynamic 3d gaussians: Tracking by persistent dynamic view synthesis,'' in \emph{3DV}, 2024.

\bibitem{yang2024deformable}
Z.~Yang, X.~Gao, W.~Zhou, S.~Jiao, Y.~Zhang, and X.~Jin, ``Deformable 3d gaussians for high-fidelity monocular dynamic scene reconstruction,'' in \emph{Proceedings of the IEEE/CVF Conference on Computer Vision and Pattern Recognition}, 2024, pp. 20\,331--20\,341.

\bibitem{li2024spacetime}
Z.~Li, Z.~Chen, Z.~Li, and Y.~Xu, ``Spacetime gaussian feature splatting for real-time dynamic view synthesis,'' in \emph{Proceedings of the IEEE/CVF Conference on Computer Vision and Pattern Recognition}, 2024, pp. 8508--8520.

\bibitem{wu20244d}
G.~Wu, T.~Yi, J.~Fang, L.~Xie, X.~Zhang, W.~Wei, W.~Liu, Q.~Tian, and X.~Wang, ``4d gaussian splatting for real-time dynamic scene rendering,'' in \emph{Proceedings of the IEEE/CVF Conference on Computer Vision and Pattern Recognition}, 2024, pp. 20\,310--20\,320.

\bibitem{marougkas2024personalized}
A.~Marougkas, C.~Troussas, A.~Krouska, and C.~Sgouropoulou, ``How personalized and effective is immersive virtual reality in education? a systematic literature review for the last decade,'' \emph{Multimedia Tools and Applications}, vol.~83, no.~6, pp. 18\,185--18\,233, 2024.

\bibitem{behley2019iccv}
J.~Behley, M.~Garbade, A.~Milioto, J.~Quenzel, S.~Behnke, C.~Stachniss, and J.~Gall, ``{SemanticKITTI: A Dataset for Semantic Scene Understanding of LiDAR Sequences},'' in \emph{Proc. of the IEEE/CVF International Conf.~on Computer Vision (ICCV)}, 2019.

\bibitem{liao2022kitti}
Y.~Liao, J.~Xie, and A.~Geiger, ``Kitti-360: A novel dataset and benchmarks for urban scene understanding in 2d and 3d,'' \emph{IEEE Transactions on Pattern Analysis and Machine Intelligence}, vol.~45, no.~3, pp. 3292--3310, 2022.

\bibitem{ben2021ikea}
Y.~Ben-Shabat, X.~Yu, F.~Saleh, D.~Campbell, C.~Rodriguez-Opazo, H.~Li, and S.~Gould, ``The ikea asm dataset: Understanding people assembling furniture through actions, objects and pose,'' in \emph{Proceedings of the IEEE/CVF Winter Conference on Applications of Computer Vision}, 2021, pp. 847--859.

\bibitem{yin2023hi4d}
Y.~Yin, C.~Guo, M.~Kaufmann, J.~J. Zarate, J.~Song, and O.~Hilliges, ``Hi4d: 4d instance segmentation of close human interaction,'' in \emph{Proceedings of the IEEE/CVF Conference on Computer Vision and Pattern Recognition}, 2023, pp. 17\,016--17\,027.

\bibitem{mustafa20174d}
A.~Mustafa, M.~Volino, J.-Y. Guillemaut, and A.~Hilton, ``4d temporally coherent light-field video,'' in \emph{2017 International Conference on 3D Vision (3DV)}.\hskip 1em plus 0.5em minus 0.4em\relax IEEE, 2017, pp. 29--37.

\bibitem{ionescu2013human3}
C.~Ionescu, D.~Papava, V.~Olaru, and C.~Sminchisescu, ``Human3. 6m: Large scale datasets and predictive methods for 3d human sensing in natural environments,'' \emph{IEEE transactions on pattern analysis and machine intelligence}, vol.~36, no.~7, pp. 1325--1339, 2013.

\bibitem{kim2012outdoor}
H.~Kim, J.-Y. Guillemaut, T.~Takai, M.~Sarim, and A.~Hilton, ``Outdoor dynamic 3-d scene reconstruction,'' \emph{IEEE Transactions on Circuits and Systems for Video Technology}, vol.~22, no.~11, pp. 1611--1622, 2012.

\bibitem{zitnick2004high}
C.~L. Zitnick, S.~B. Kang, M.~Uyttendaele, S.~Winder, and R.~Szeliski, ``High-quality video view interpolation using a layered representation,'' \emph{ACM transactions on graphics (TOG)}, vol.~23, no.~3, pp. 600--608, 2004.

\bibitem{ballan2010unstructured}
L.~Ballan, G.~J. Brostow, J.~Puwein, and M.~Pollefeys, ``Unstructured video-based rendering: Interactive exploration of casually captured videos,'' in \emph{ACM SIGGRAPH 2010 papers}, 2010, pp. 1--11.

\bibitem{ding2023mose}
H.~Ding, C.~Liu, S.~He, X.~Jiang, P.~H. Torr, and S.~Bai, ``Mose: A new dataset for video object segmentation in complex scenes,'' in \emph{Proceedings of the IEEE/CVF International Conference on Computer Vision}, 2023, pp. 20\,224--20\,234.

\bibitem{kim2020video}
D.~Kim, S.~Woo, J.-Y. Lee, and I.~S. Kweon, ``Video panoptic segmentation,'' in \emph{Proceedings of the IEEE/CVF Conference on Computer Vision and Pattern Recognition}, 2020, pp. 9859--9868.

\bibitem{barron2022mip}
J.~T. Barron, B.~Mildenhall, D.~Verbin, P.~P. Srinivasan, and P.~Hedman, ``Mip-nerf 360: Unbounded anti-aliased neural radiance fields,'' in \emph{Proceedings of the IEEE/CVF Conference on Computer Vision and Pattern Recognition}, 2022, pp. 5470--5479.

\bibitem{chen2022tensorf}
A.~Chen, Z.~Xu, A.~Geiger, J.~Yu, and H.~Su, ``Tensorf: Tensorial radiance fields,'' in \emph{European Conference on Computer Vision}.\hskip 1em plus 0.5em minus 0.4em\relax Springer, 2022, pp. 333--350.

\bibitem{kerbl20233d}
B.~Kerbl, G.~Kopanas, T.~Leimk{\"u}hler, and G.~Drettakis, ``3d gaussian splatting for real-time radiance field rendering,'' \emph{ACM Transactions on Graphics (ToG)}, vol.~42, no.~4, pp. 1--14, 2023.

\bibitem{garfield2024}
C.~M. Kim, M.~Wu, J.~Kerr, M.~Tancik, K.~Goldberg, and A.~Kanazawa, ``Garfield: Group anything with radiance fields,'' in \emph{arXiv}, 2024.

\bibitem{NeRFshop23}
\BIBentryALTinterwordspacing
C.~Jambon, B.~Kerbl, G.~Kopanas, S.~Diolatzis, T.~Leimk{\"u}hler, and G.~Drettakis, ``Nerfshop: Interactive editing of neural radiance fields,'' \emph{Proceedings of the ACM on Computer Graphics and Interactive Techniques}, vol.~6, no.~1, May 2023. [Online]. Available: \url{https://repo-sam.inria.fr/fungraph/nerfshop/}
\BIBentrySTDinterwordspacing

\bibitem{zhi2021place}
S.~Zhi, T.~Laidlow, S.~Leutenegger, and A.~J. Davison, ``In-place scene labelling and understanding with implicit scene representation,'' in \emph{Proceedings of the IEEE/CVF International Conference on Computer Vision}, 2021, pp. 15\,838--15\,847.

\bibitem{siddiqui2023panoptic}
Y.~Siddiqui, L.~Porzi, S.~R. Bul{\`o}, N.~M{\"u}ller, M.~Nie{\ss}ner, A.~Dai, and P.~Kontschieder, ``Panoptic lifting for 3d scene understanding with neural fields,'' in \emph{Proceedings of the IEEE/CVF Conference on Computer Vision and Pattern Recognition}, 2023, pp. 9043--9052.

\bibitem{kundu2022panoptic}
A.~Kundu, K.~Genova, X.~Yin, A.~Fathi, C.~Pantofaru, L.~J. Guibas, A.~Tagliasacchi, F.~Dellaert, and T.~Funkhouser, ``Panoptic neural fields: A semantic object-aware neural scene representation,'' in \emph{Proceedings of the IEEE/CVF Conference on Computer Vision and Pattern Recognition}, 2022, pp. 12\,871--12\,881.

\bibitem{straub2019replica}
J.~Straub, T.~Whelan, L.~Ma, Y.~Chen, E.~Wijmans, S.~Green, J.~J. Engel, R.~Mur-Artal, C.~Ren, S.~Verma \emph{et~al.}, ``The replica dataset: A digital replica of indoor spaces,'' \emph{arXiv preprint arXiv:1906.05797}, 2019.

\bibitem{roberts2021hypersim}
M.~Roberts, J.~Ramapuram, A.~Ranjan, A.~Kumar, M.~A. Bautista, N.~Paczan, R.~Webb, and J.~M. Susskind, ``Hypersim: A photorealistic synthetic dataset for holistic indoor scene understanding,'' in \emph{Proceedings of the IEEE/CVF international conference on computer vision}, 2021, pp. 10\,912--10\,922.

\bibitem{geiger2012we}
A.~Geiger, P.~Lenz, and R.~Urtasun, ``Are we ready for autonomous driving? the kitti vision benchmark suite,'' in \emph{2012 IEEE conference on computer vision and pattern recognition}.\hskip 1em plus 0.5em minus 0.4em\relax IEEE, 2012, pp. 3354--3361.

\bibitem{ren2022neural}
Z.~Ren, A.~Agarwala, B.~Russell, A.~G. Schwing, and O.~Wang, ``Neural volumetric object selection,'' in \emph{Proceedings of the IEEE/CVF Conference on Computer Vision and Pattern Recognition}, 2022, pp. 6133--6142.

\bibitem{mildenhall2019local}
B.~Mildenhall, P.~P. Srinivasan, R.~Ortiz-Cayon, N.~K. Kalantari, R.~Ramamoorthi, R.~Ng, and A.~Kar, ``Local light field fusion: Practical view synthesis with prescriptive sampling guidelines,'' \emph{ACM Transactions on Graphics (TOG)}, vol.~38, no.~4, pp. 1--14, 2019.

\bibitem{mirzaei2023spin}
A.~Mirzaei, T.~Aumentado-Armstrong, K.~G. Derpanis, J.~Kelly, M.~A. Brubaker, I.~Gilitschenski, and A.~Levinshtein, ``Spin-nerf: Multiview segmentation and perceptual inpainting with neural radiance fields,'' in \emph{Proceedings of the IEEE/CVF Conference on Computer Vision and Pattern Recognition}, 2023, pp. 20\,669--20\,679.

\bibitem{wang2019fast}
Q.~Wang, L.~Zhang, L.~Bertinetto, W.~Hu, and P.~H. Torr, ``Fast online object tracking and segmentation: A unifying approach,'' in \emph{Proceedings of the IEEE/CVF conference on Computer Vision and Pattern Recognition}, 2019, pp. 1328--1338.

\bibitem{chen2020state}
X.~Chen, Z.~Li, Y.~Yuan, G.~Yu, J.~Shen, and D.~Qi, ``State-aware tracker for real-time video object segmentation,'' in \emph{Proceedings of the IEEE/CVF conference on computer vision and pattern recognition}, 2020, pp. 9384--9393.

\bibitem{cheng2022xmem}
H.~K. Cheng and A.~G. Schwing, ``Xmem: Long-term video object segmentation with an atkinson-shiffrin memory model,'' in \emph{European Conference on Computer Vision}.\hskip 1em plus 0.5em minus 0.4em\relax Springer, 2022, pp. 640--658.

\bibitem{gadde2017semantic}
R.~Gadde, V.~Jampani, and P.~V. Gehler, ``Semantic video cnns through representation warping,'' in \emph{Proceedings of the IEEE International Conference on Computer Vision}, 2017, pp. 4453--4462.

\bibitem{yang2019video}
L.~Yang, Y.~Fan, and N.~Xu, ``Video instance segmentation,'' in \emph{Proceedings of the IEEE/CVF International Conference on Computer Vision}, 2019, pp. 5188--5197.

\bibitem{qiao2021vip}
S.~Qiao, Y.~Zhu, H.~Adam, A.~Yuille, and L.-C. Chen, ``Vip-deeplab: Learning visual perception with depth-aware video panoptic segmentation,'' in \emph{Proceedings of the IEEE/CVF Conference on Computer Vision and Pattern Recognition}, 2021, pp. 3997--4008.

\bibitem{wang2021amanet}
X.~Wang, Y.~Guo, J.~Song, L.~Gao, and H.~T. Shen, ``Amanet: Adaptive multi-path aggregation for learning human 2d-3d correspondences,'' \emph{IEEE Transactions on Multimedia}, vol.~25, pp. 979--992, 2021.

\bibitem{wang2024cpi}
X.~Wang, X.~Chen, L.~Gao, J.~Song, and H.~T. Shen, ``Cpi-parser: Integrating causal properties into multiple human parsing,'' \emph{IEEE Transactions on Image Processing}, 2024.

\bibitem{yoon2020novel}
J.~S. Yoon, K.~Kim, O.~Gallo, H.~S. Park, and J.~Kautz, ``Novel view synthesis of dynamic scenes with globally coherent depths from a monocular camera,'' in \emph{Proceedings of the IEEE/CVF Conference on Computer Vision and Pattern Recognition}, 2020, pp. 5336--5345.

\bibitem{luo2020consistent}
X.~Luo, J.-B. Huang, R.~Szeliski, K.~Matzen, and J.~Kopf, ``Consistent video depth estimation,'' \emph{ACM Transactions on Graphics (ToG)}, vol.~39, no.~4, pp. 71--1, 2020.

\bibitem{peng2021neural}
S.~Peng, Y.~Zhang, Y.~Xu, Q.~Wang, Q.~Shuai, H.~Bao, and X.~Zhou, ``Neural body: Implicit neural representations with structured latent codes for novel view synthesis of dynamic humans,'' in \emph{CVPR}, 2021.

\bibitem{fang2021mirrored}
Q.~Fang, Q.~Shuai, J.~Dong, H.~Bao, and X.~Zhou, ``Reconstructing 3d human pose by watching humans in the mirror,'' in \emph{CVPR}, 2021.

\bibitem{sigal2010humaneva}
L.~Sigal, A.~O. Balan, and M.~J. Black, ``Humaneva: Synchronized video and motion capture dataset and baseline algorithm for evaluation of articulated human motion,'' \emph{International journal of computer vision}, vol.~87, no. 1-2, pp. 4--27, 2010.

\bibitem{mustafa2020semantically}
A.~Mustafa and A.~Hilton, ``Semantically coherent 4d scene flow of dynamic scenes,'' \emph{International Journal of Computer Vision}, vol. 128, no.~2, pp. 319--335, 2020.

\bibitem{doyen2017light}
D.~Doyen, T.~Langlois, B.~Vandame, F.~Babon, G.~Boisson, N.~Sabater, R.~Gendrot, and A.~Schubert, ``Light field content from 16-camera rig,'' \emph{ISO/IEC JTC1/SC29 WG11 Doc. m40010. Geneva}, 2017.

\bibitem{tapie2021barn}
T.~Tapie, A.~Schubert, R.~Gendrot, G.~Briand, F.~Thudor, and R.~Doré, ``Barn new natural content proposal for miv,'' ISO/IEC JTC 1/SC 29/WG 4 m56632, Online, April 2021.

\bibitem{tapie2021breakfast}
------, ``Breakfast new natural content proposal for miv,'' ISO/IEC JTC 1/SC 29/WG 4 m56730, Online, April 2021.

\bibitem{salahieh2018kermit}
B.~Salahieh, B.~Marva, M.~Nentedem, A.~Kumar, V.~Popvic, K.~Seshadrinathan, O.~Nestares, and J.~Boyce, ``Kermit test sequence for windowed 6dof activities,'' \emph{ISO/IEC JTC1/SC29/WG11 MPEG M43748}, July 2018.

\bibitem{mieloch2020mpeg}
D.~Mieloch, A.~Dziembowski, and M.~Doma{\'n}ski, ``[{MPEG-I Visual}] {Natural outdoor test sequences},'' \emph{ISO/IEC JTC1/SC29/WG11 MPEG2020/M51598}, January 2020.

\bibitem{domanski2016multiview}
M.~Doma{\'n}ski, A.~Dziembowski, A.~Grzelka, D.~Mieloch, O.~Stankiewicz, and K.~Wegner, ``Multiview test video sequences for free navigation exploration obtained using pairs of cameras,'' \emph{ISO/IEC JTC1/SC29/WG11, Doc. MPEG M}, vol. 38247, 2016.

\bibitem{sheng2021new}
X.~Sheng, ``New test sequence for mpeg-i visual,'' ISO/IEC JTC1/SC29/WG04 MPEG136/M58275, Online, October 2021.

\bibitem{mieloch2023new}
D.~Mieloch, A.~Dziembowski, B.~Szydełko, D.~Klóska, A.~Grzelka, J.~Stankowski, M.~Domański, G.~Lee, and J.~Jeong, ``[{MIV}] new natural content - martialarts,'' ISO/IEC JTC1/SC29/WG4 MPEG 141, M61949, January 2023.

\bibitem{li2022neural}
T.~Li, M.~Slavcheva, M.~Zollhoefer, S.~Green, C.~Lassner, C.~Kim, T.~Schmidt, S.~Lovegrove, M.~Goesele, R.~Newcombe \emph{et~al.}, ``Neural 3d video synthesis from multi-view video,'' in \emph{Proceedings of the IEEE/CVF Conference on Computer Vision and Pattern Recognition}, 2022, pp. 5521--5531.

\bibitem{broxton2020immersive}
M.~Broxton, J.~Flynn, R.~Overbeck, D.~Erickson, P.~Hedman, M.~Duvall, J.~Dourgarian, J.~Busch, M.~Whalen, and P.~Debevec, ``Immersive light field video with a layered mesh representation,'' \emph{ACM Transactions on Graphics (TOG)}, vol.~39, no.~4, pp. 86--1, 2020.

\bibitem{hobloss2021multi}
N.~Hobloss, L.~Zhang, and M.~Cagnazzo, ``A multi-view stereoscopic video database with green screen (mtf) for video transition quality-of-experience assessment,'' in \emph{2021 13th International Conference on Quality of Multimedia Experience (QoMEX)}.\hskip 1em plus 0.5em minus 0.4em\relax IEEE, 2021, pp. 201--206.

\bibitem{domanski2014poznan}
M.~Domanski, A.~Dziembowski, A.~Kuehn, M.~Kurc, A.~Luczak, D.~Mieloch, J.~Siast, O.~Stankiewicz, and K.~Wegner, ``Poznan blocks-a multiview video test sequence and camera parameters for free viewpoint television,'' \emph{ISO/IEC JTC1/SC29/WG11 MPEG2014 M}, vol. 32243, pp. 13--17, 2014.

\bibitem{miao2022large}
J.~Miao, X.~Wang, Y.~Wu, W.~Li, X.~Zhang, Y.~Wei, and Y.~Yang, ``Large-scale video panoptic segmentation in the wild: A benchmark,'' in \emph{Proceedings of the IEEE/CVF Conference on Computer Vision and Pattern Recognition}, 2022, pp. 21\,033--21\,043.

\bibitem{kirillov2023segment}
A.~Kirillov, E.~Mintun, N.~Ravi, H.~Mao, C.~Rolland, L.~Gustafson, T.~Xiao, S.~Whitehead, A.~C. Berg, W.-Y. Lo \emph{et~al.}, ``Segment anything,'' \emph{arXiv preprint arXiv:2304.02643}, 2023.

\bibitem{zamani2017similarity}
Y.~Zamani, H.~Shirzad, and S.~Kasaei, ``Similarity measures for intersection of camera view frustums,'' in \emph{2017 10th Iranian Conference on Machine Vision and Image Processing (MVIP)}.\hskip 1em plus 0.5em minus 0.4em\relax IEEE, 2017, pp. 171--175.

\bibitem{lindenberger2023lightglue}
P.~Lindenberger, P.-E. Sarlin, and M.~Pollefeys, ``Lightglue: Local feature matching at light speed,'' \emph{arXiv preprint arXiv:2306.13643}, 2023.

\bibitem{tang2023scene}
S.~Tang, W.~Pei, X.~Tao, T.~Jia, G.~Lu, and Y.-W. Tai, ``Scene-generalizable interactive segmentation of radiance fields,'' in \emph{Proceedings of the 31st ACM International Conference on Multimedia}, 2023, pp. 6744--6755.

\bibitem{cheng2023tracking}
H.~K. Cheng, S.~W. Oh, B.~Price, A.~Schwing, and J.-Y. Lee, ``Tracking anything with decoupled video segmentation,'' in \emph{Proceedings of the IEEE/CVF International Conference on Computer Vision}, 2023, pp. 1316--1326.

\bibitem{caron2021emerging}
M.~Caron, H.~Touvron, I.~Misra, H.~J{\'e}gou, J.~Mairal, P.~Bojanowski, and A.~Joulin, ``Emerging properties in self-supervised vision transformers,'' in \emph{Proceedings of the IEEE/CVF international conference on computer vision}, 2021, pp. 9650--9660.

\end{thebibliography}

\end{document}